\documentclass[letterpaper]{article} 
\usepackage{aaai2026}  
\usepackage{times}  
\usepackage{helvet}  
\usepackage{courier}  
\usepackage[hyphens]{url}  
\usepackage{graphicx} 
\usepackage{natbib}  
\usepackage{caption} 
\usepackage{multirow}
\usepackage{booktabs}
\usepackage[caption=false,font=footnotesize,labelfont=sf,textfont=sf]{subfig}
\usepackage{amsmath} 
\usepackage{algorithm}
\usepackage{algorithmic}
\usepackage{pifont}
\usepackage{comment}
\usepackage{newfloat}
\usepackage{listings}
\usepackage[T1]{fontenc}
\usepackage[polish,english]{babel}

\DeclareCaptionStyle{ruled}{labelfont=normalfont,labelsep=colon,strut=off} 
\floatstyle{ruled}
\newfloat{listing}{tb}{lst}{}
\floatname{listing}{Listing}
\newif\ifappendixincluded
\appendixincludedtrue

\title{On Stealing Graph Neural Network Models}
\author {
    Marcin Podhajski\textsuperscript{\rm 1,2}\thanks{Corresponding author: marcin.podhajski@ideas-ncbr.pl},
    Jan Dubiński\textsuperscript{\rm 3,4},
    Franziska Boenisch\textsuperscript{\rm 5},
    Adam Dziedzic\textsuperscript{\rm 5},
    Agnieszka Pr\k{e}gowska\textsuperscript{\rm 1},
    Tomasz P. Michalak\textsuperscript{\rm 6,7}
}
\affiliations {
    \textsuperscript{\rm 1}Institute of Fundamental Technological Research, Polish Academy of Sciences,
    \textsuperscript{\rm 2}IDEAS NCBR,
    \textsuperscript{\rm 3}Warsaw University of Technology,
    \textsuperscript{\rm 4}NASK National Research Institute,
    \textsuperscript{\rm 5}CISPA Helmholtz Center for Information Security,
    \textsuperscript{\rm 6}University of Warsaw,
    \textsuperscript{\rm 7}IDEAS Research Institute\\
}

\newcommand{\reb}[1]{\textcolor{black}{#1}}

\begin{document}

\maketitle

\begin{abstract}

Current graph neural network (GNN) model-stealing methods rely heavily on queries to the victim model, assuming no hard query limits. However, in reality, the number of allowed queries can be severely limited. In this paper, we demonstrate how an adversary can extract a GNN with very limited interactions with the model.  Our approach first enables the adversary to obtain the model backbone without making direct queries to the victim model and then to strategically utilize a fixed query limit to extract the most informative data. 
The experiments on eight real-world datasets demonstrate the effectiveness of the attack, even under a very restricted query limit and under defense against model extraction in place. Our findings underscore the need for robust defenses against GNN model extraction threats.
\end{abstract}

\begin{links}
    \link{Code}{https://github.com/m-podhajski/OnStealingGNNs}
    \link{Extended version}{https://arxiv.org/pdf/2511.07170}
\end{links}

\section{Introduction}
\par
Graph Neural Networks have recently become the center of attention in many dynamically developing Artificial Intelligence-based technologies. They have been successfully applied to various tasks involving graph-structured data: node classification, link prediction, graph classification, and recommendation systems \cite{Sharma2024}. Unfortunately, like all types of neural networks, GNNs are vulnerable to various security threats, including adversarial attacks \cite{adversarial}, data poisoning \cite{poisoning}, model inversion \cite{inversion}, and privacy breaches \cite{Guan2024, membershipGNN}. 

\reb{Particularly, GNN models are vulnerable to \emph{model-stealing attacks}~\cite{TZJRR16,JCBKP20, pmlr-v162-dziedzic22a}. In such attacks, an adversary with query access to a target (victim) model can replicate its functionality by training a local surrogate model on query-response pairs. A typical defense against such an attack is to limit the number of queries. However, previous studies on GNN model stealing \cite{Podhajski_2024, shen2021model, datafree, wu2021model} generally assume access to a relatively large number of queries, focusing on maximizing performance as the budget increases. This overlooks the practical reality that in many applications, adversaries must operate under severe query restrictions.}

\begin{figure}[t]
\begin{center}

    \includegraphics[scale=0.341]{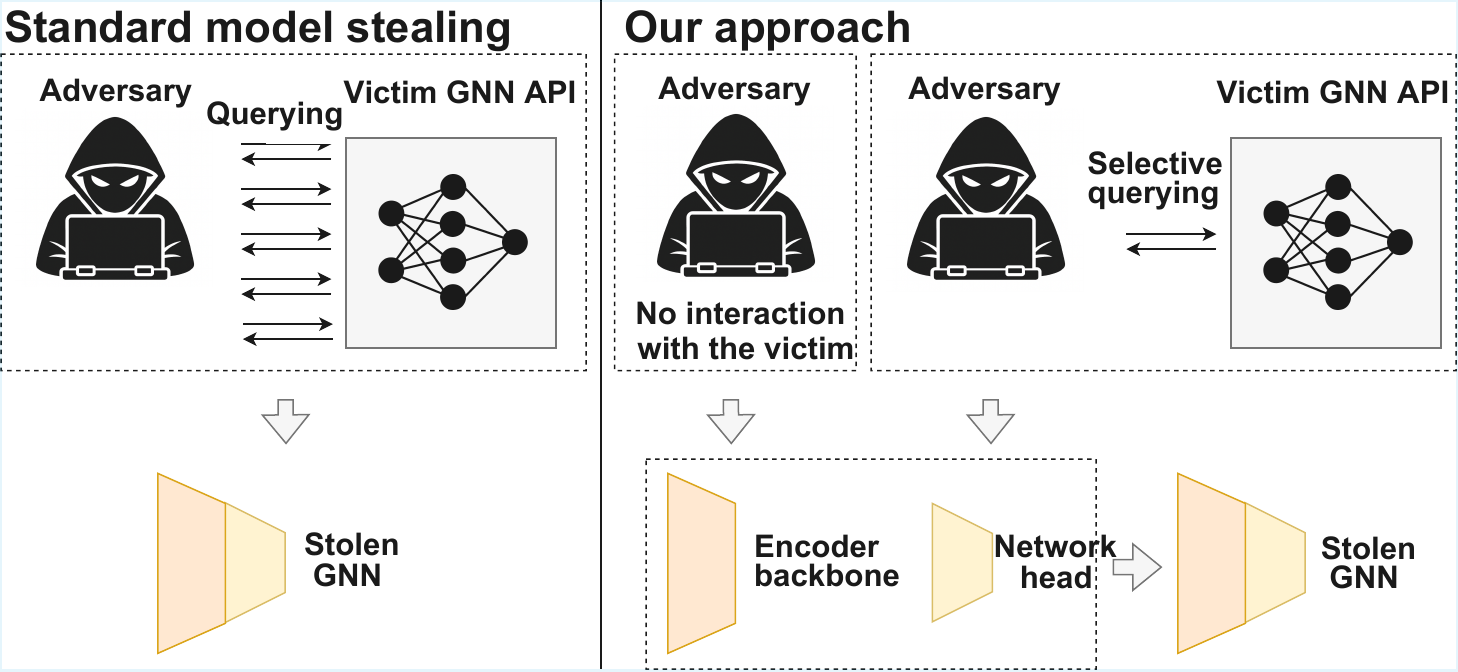} 
\end{center}

\caption{\textbf{Standard GNN model stealing vs. our approach.} Conventional model stealing methods extract the entire GNN through extensive querying of the victim model API. Our method divides this process into stages, focusing on maximizing the stealing outcome within a restricted query limit. First, we show that the adversary can obtain the encoder backbone locally, without any interaction with the victim API. Then, the adversary performs query selection using the representations from the extracted encoder and extracts the network head via selective querying. This enables effective model stealing under strict query budgets, demonstrating that the GNN model stealing threat is significantly more severe than previously assumed.} 
\label{fig:teaser}%
\end{figure}

\reb{Our method challenges this conventional approach (summarized visually in Figure \ref{fig:teaser}) and shows that a GNN model can be effectively extracted even when the adversary faces strict constraints on the number of queries allowed to the victim model. It first enables the adversary to recover the model backbone without querying the victim directly, then strategically uses a fixed query budget to extract the most informative data for effective model stealing.}

Collectively, our results demonstrate that model-stealing attacks on GNNs are both more effective and resource-efficient than previously understood, highlighting the critical severity of these attacks in both inductive and transductive scenarios. For instance, targeting a SAGE \cite{hamilton2017inductive} model trained on the Physics dataset, we achieve 91\% accuracy with only 100 queries to the victim model --— compared to approximately 5,000 queries, ${\sim}15\times$ higher computational cost, and additional victim output (such as embeddings) needed by the current state-of-the-art method to reach similar accuracy. 

In summary, our contributions are as follows:
\begin{itemize}
\item We identify a threat that has been previously overlooked in research: the ability to steal a GNN model even under a significant limit on access to the victim.
\item We investigate this threat by showing that an adversary with access to query data, but no direct access to the model, can locally obtain a high-quality encoder at a low computational cost.
\item We further demonstrate that an adversary with restricted access to the victim model can strategically select queries to train the model head, resulting in a stolen model with improved accuracy and fidelity.
\end{itemize}

\section{Background}
We first introduce the basic notions considered in this work and move to the motivation behind our method.

\subsection{Graph Neural Networks}

Graph Neural Networks are a class of neural architectures that use the graph structure $\mathbf{A}$ and node features $\mathbf{X}$ as inputs. They are utilized across tasks such as node classification ~\cite{ KW17}, link prediction ~\cite{PAS14, GL16}, and graph/subgraph classification ~\cite{NEURIPS2020_5bca8566}.

Our work addresses model-stealing vulnerabilities in GNNs, specifically for node-level tasks. For such tasks, GNNs are typically trained and evaluated in two scenarios crucial to understanding these vulnerabilities: transductive and inductive settings. In the transductive paradigm, the data consists of a fixed graph with a portion of nodes labeled for training, while another portion remains unlabeled. The objective is to apply these labeled nodes to predict labels for the previously unlabeled ones within the same graph. However, transductive models are limited in their ability to generalize to novel nodes. Conversely, in the inductive setting, GNNs are designed to expand their learning to accommodate unseen nodes or graphs that were not part of the training data. 

\subsection{Model Stealing Attacks}
\label{sec:extaction-attacs}
\reb{Model stealing attacks aim to replicate the functionality of a victim model $f_v$ trained on a private dataset $\mathbf{G}_V$. An attacker with black-box access to the victim model first selects a query graph $\mathbf{G}_Q$ from their available data. For each query node $v_i$ from $\mathbf{G}_Q$, the attacker obtains the corresponding output $y_i = f_v(v_i)$ from the victim model. The attacker then constructs a surrogate training dataset, $\{(v_i, y_i)\}$, which is then used to train a surrogate model $f_s$ that mimics the behavior of $f_v$.}
 Model extraction attacks have been demonstrated to be successful against various types of models, including classifiers \cite{JCBKP20,TZJRR16} and encoders \cite{pmlr-v162-dziedzic22a,ContSteal}.

\begin{table}[ht!]
    \caption{Existing stealing attacks on node-level GNNs.}
\small
    \centering
    
       \setlength{\tabcolsep}{1pt}
    \begin{tabular}{lccccc}
        \toprule
         & Trans. & Ind. & Data Acc. & Vict. Q. Acc. & Vict. Output \\
         \midrule
        \citeauthor{datafree} & \ding{51} & \ding{51} & None & Unlimited & Only Pred. \\
        \citeauthor{wu2021model} & \ding{51} & \ding{55} & Limited & Unlimited & Only Pred. \\
        \citeauthor{Podhajski_2024} & \ding{55} & \ding{51} & Limited & Unlimited & Emb. \& Pred. \\
        \citeauthor{shen2021model} & \ding{55} & \ding{51} & Limited & Unlimited & Emb. \& Pred. \\
        \textbf{Ours} & \ding{51} & \ding{51} & Limited & Limited & Only Pred.\\
        \bottomrule
    \end{tabular}
    \label{tab:methods}
\end{table}

 \textbf{Relation to existing work.} 
 An adversary attempting to steal a model faces two primary challenges: limited data access or restricted access to the model itself. Existing approaches to GNN model stealing predominantly address scenarios with limited or no data access, but none explore the problem of severely restricted model access. Specifically, \citeauthor{shen2021model} and \citeauthor{Podhajski_2024} concentrate on stealing the encoder in the inductive setting, assuming unlimited access to the model but limited access to data. \reb{Both of these methods leverage the victim model's output (e.g., embeddings) to train the encoder and then train Multi-Layer Perceptrons (MLPs) on top of the frozen encoder using available class labels. This requirement for intermediate embeddings contrasts with our work, which assumes a more restrictive and practical threat model where adversaries only receive the final class predictions.} Similarly, \citeauthor{wu2021model} focuses on limited data access in the transductive setting, while also assuming unrestricted access to the victim. Another contribution is the data-free model extraction attack framework proposed by \citeauthor{datafree}. This framework enables GNN model extraction without requiring access to actual node features or graph structures. However, it assumes unlimited access to the victim model and relies on an extensive number of queries.

A comprehensive comparison of various GNN model-stealing methods, including the assumptions regarding data and model access, is presented in Table \ref{tab:methods}, highlighting the diverse methodologies and their limitations.

\subsection{Self-Supervised Learning in GNNs}
Self-supervised learning (SSL) for GNNs trains models on data without explicit labels, focusing on node or graph representations. Representative methods include Deep Graph Infomax (DGI) by \citet{velickovic2018deep}, Latent Graph Prediction (LaGraph) by \citet{pmlr-v162-xie22e}, and Bootstrapped Graph Latents (BGRL) by \citet{thakoor2022large}.

In the context of model-stealing attacks, the capability of learning useful feature representations from unlabeled data may be crucial when the adversary has access to surplus data, \textit{i.e.}, additional data points beyond the victim model query limit. Thus, SSL techniques can leverage this surplus data to refine and enhance the stolen model’s performance. 
This means that SSL can be a powerful tool for overcoming constraints imposed on traditional model-stealing methods. SSL allows adversaries to bypass limitations imposed by direct querying of the victim model. By training on unlabeled data and utilizing self-supervised objectives, SSL can produce feature representations that are robust and informative, even without extensive interaction with the victim model. This is especially beneficial when direct access to the victim model is limited or restricted.

\subsection{Motivation}

\begin{table}[h]
\centering
\small
\setlength{\tabcolsep}{2pt}
\caption{Comparison of test accuracies using a randomly initialized GCN encoder with a trained MLP head vs.\ a fully SSL-trained GCN encoder in both inductive and transductive settings, as reported by DGI~\cite{velickovic2018deep} and BGRL~\cite{thakoor2022large}.}
\label{tab:ssl_results}
\begin{tabular}{lccccc}
\toprule
\textbf{Setting} & \textbf{Dataset} & \textbf{Random} & \textbf{SSL-Trained} & \textbf{Gain} & \textbf{Method} \\
\midrule
\multirow{7}{*}{Inductive} 
& Reddit     & 93.3 & 94.0 & 0.7 & DGI \\ 
& PPI        & 62.6 & 63.8 & 1.2 & DGI \\ 
& WikiCS     & 78.9 & 79.9 & 1.0 & BGRL \\
& Computer   & 86.5 & 90.3 & 3.8 & BGRL \\
& Photo      & 92.0 & 93.1 & 1.1 & BGRL \\
& CS         & 91.6 & 93.3 & 1.7 & BGRL \\
& Physics    & 93.7 & 95.7 & 2.0 & BGRL \\
\midrule
\multirow{3}{*}{Transductive} 
& Cora       & 69.3 & 82.3 & 13.0 & DGI \\ 
& Citeseer   & 61.9 & 71.8 & 9.9  & DGI \\ 
& Pubmed     & 69.6 & 76.8 & 7.2  & DGI \\ 
\bottomrule
\end{tabular}
\end{table}

\reb{Existing works on inductive GNN model stealing, such as those by \citeauthor{shen2021model} and \citeauthor{Podhajski_2024}, rely on stealing the victim's encoder using rich responses like query embeddings. However, a key observation from self-supervised learning provides a more efficient, query-free path for the adversary. }
In the inductive graph learning setting, \citet{velickovic2018deep} showed that a randomly initialized graph convolutional network (GCN) can already extract informative features and serve as a strong baseline. This phenomenon is linked to the Weisfeiler-Lehman graph isomorphism test~\cite{weisfeiler1968reduction}, as discussed by \citet{KW17, hamilton2017inductive}.
As shown by recent works, a randomly initialized GNN encoder, when paired with a trained MLP head, can achieve performance comparable to that of a fully trained model. For instance, DGI~\cite{velickovic2018deep} reports test accuracies of 93.3\% (random encoder) vs. 94.0\% (SSL-trained) on Reddit, and 62.6\% vs. 63.8\% on PPI. Similarly, BGRL~\cite{thakoor2022large} observes minimal performance gaps across datasets such as WikiCS, Computer, Photo, CS, and Physics. These results are summarized in Table~\ref{tab:ssl_results}. 
Our experimental findings align with these insights. T-SNE visualizations (\ifappendixincluded
Figure~\ref {fig:tsne_combined} in the Appendix\fi) suggest that even randomly initialized encoders produce structured embeddings in the inductive setting.

In contrast, the transductive setting sees greater benefits from self-supervised learning. As shown in Table~\ref{tab:ssl_results}, training a GNN encoder with SSL yields notable performance gains. Since the full graph structure (including test node connectivity) is available during training, the encoder can learn representations that better align with the downstream task -- unlike in the inductive setting, where such structural information is unavailable~\cite{velickovic2018deep, thakoor2022large}. These improvements are further supported by the visualizations in
\ifappendixincluded
Figure~\ref{fig:tsne_combined} \reb{in the Appendix}\fi. Notably, transductive graphs are typically small, keeping SSL training costs low and making it a viable option even with limited computational resources.

In the context of model stealing, these observations suggest that in both settings, in practice, the encoder can be obtained without interacting with the victim model—either by using a randomly initialized backbone (inductive) or by training an encoder locally via SSL (transductive). We confirm this hypothesis empirically in Section~\ref{sec:empirical_encoder}.
Moreover, this approach benefits from low computational requirements. Randomly initialized encoders in the inductive setting and lightweight SSL training in the transductive case enable adversaries to operate under limited resources. This highlights the practicality and severity of model-stealing threats under realistic constraints.

With the encoder part of the model fixed, the central challenge becomes selecting informative queries to obtain the model head. We hypothesize that naive random query sampling is insufficient to expose the victim model’s decision boundaries. We address this by leveraging the encoder’s representations to guide query selection. Specifically, we apply clustering to the node embeddings and choose representative nodes near cluster centroids. This ensures coverage across the input space and increases the likelihood that each query contributes new information, as confirmed in Section~\ref{sec:empirical_queries}.

\section{Method}
\begin{figure}[t!]
\begin{center}
    \includegraphics[scale=0.36]{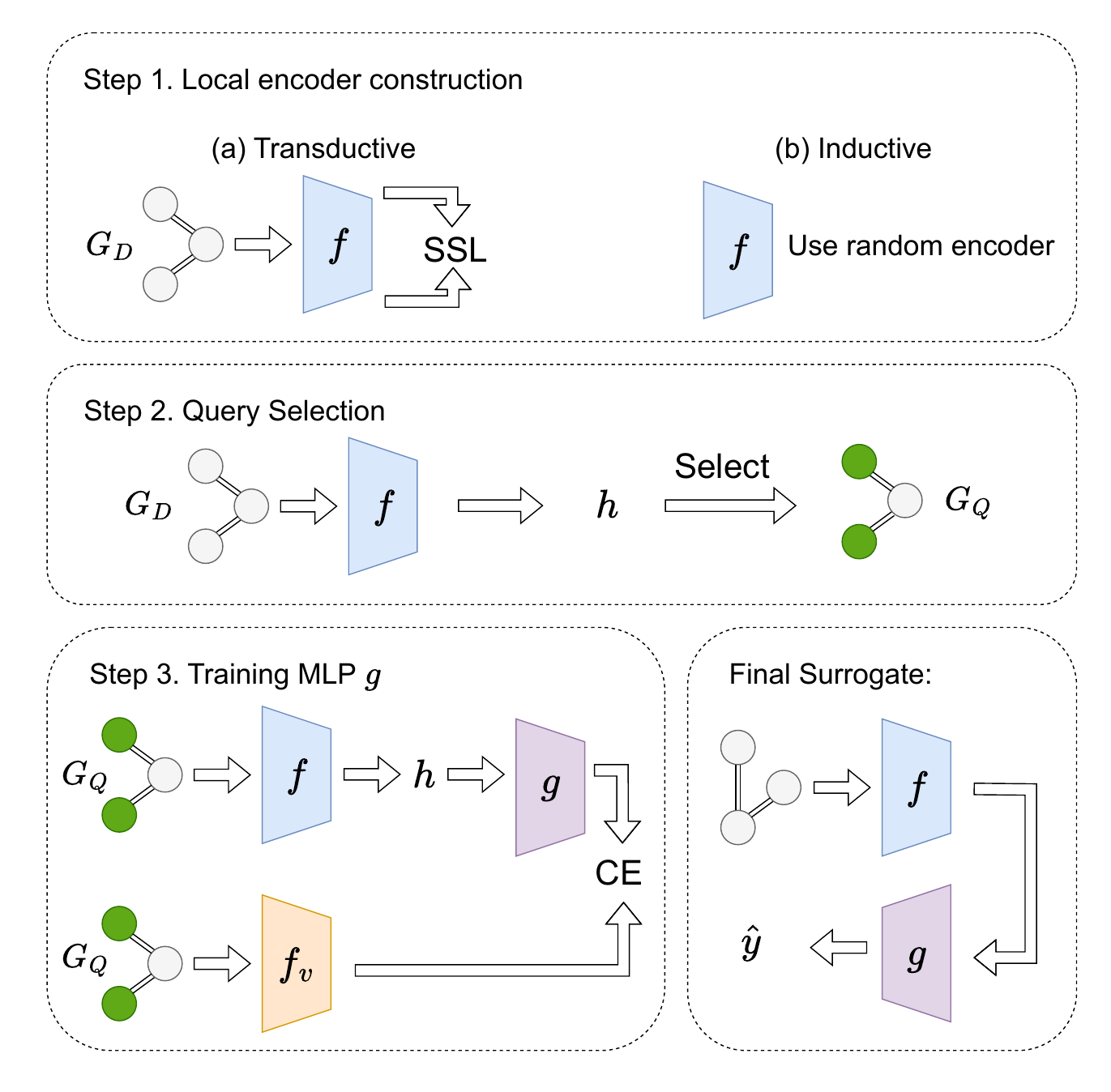} 
\end{center}

\caption{\textbf{Proposed Approach}. We first train an encoder using SSL on the adversary’s full data $G_D$ in the transductive setting, or use a random encoder in the inductive setting. Using embeddings from this encoder, we select queries $G_Q$. Finally, we train an MLP and combine it with the encoder to form the surrogate model.} 
\label{fig:scheme}%
\end{figure}

In this section, we outline our proposed approach, which is designed to extract GNN models even under restrictive query limits to the victim model. The scheme of our proposed approach is illustrated in Figure \ref{fig:scheme}. First, we acquire the encoder backbone, which serves as a feature extractor, independently of the victim model. This step involves constructing a pre-trained encoder network to generate meaningful embeddings of the input data without requiring any interaction with the victim model. The encoder backbone is crucial as it provides a robust foundation for representation learning, capturing rich semantic features from the data. Next, we focus on selecting the optimal queries to interact with the victim model. This involves leveraging the embeddings generated by the encoder to identify a set of inputs that are most informative or representative of the data distribution. Finally, we extract the model head by training an MLP on top of the encoder output. The training process utilizes class-label responses %
to the selected queries. The result is a reconstructed model head that, when combined with the encoder backbone, approximates the victim model.

\subsection{Threat Model}
To introduce the necessary research methodology, we describe the threat model, outlining the attack setting as well as the adversary's goal and capabilities. 

 \textbf{Attack Setting.}
Our research operates in a challenging \textit{black-box} scenario where the adversary has no knowledge of the target GNN model's parameters, architecture, or the training graph $\mathbf{G}_V$.
 Our investigation focuses on GNNs that produce node-level query responses, taking node $v$ as input and providing the corresponding class label.  We consider a query limit $q_n$ representing the maximum number of node predictions the adversary can obtain from the victim model.

 \textbf{Adversary's Goal.}
Referring to the taxonomy defined by \cite{JCBKP20},  adversaries' goals fall into two categories, \textit{i.e.}, \textit{theft} and \textit{reconnaissance}. The \emph{theft adversary} aims to construct a surrogate model $f_s$ matching $f_v$ on the target task~\cite{TZJRR16,PMGJCS17}, violating the intellectual property in the victim model. In contrast, the \emph{reconnaissance adversary} seeks a surrogate model $f_s$ mirroring $f_v$ across all inputs. This high-fidelity match serves as a tool for subsequent attacks, such as crafting adversarial examples without direct queries to $f_v$~\cite{PMGJCS17}.

 \textbf{Adversary's Capabilities.} First, we assume that the adversary has access to a graph $\mathbf{G}_D$ representing their own available (unlabeled) dataset. Next, we assume that the adversary queries a target model $f_v$ hidden behind a publicly accessible API~\cite{TZJRR16,OSF19,HJBGZ21,HWWBSZ21}, receiving responses $\mathbf{R}$ based on an input query graph $\mathbf{G}_Q$, which is a subgraph of $\mathbf{G}_D$. The response $\mathbf{R}$ has a size of at most $q_n$. We consider only class labels as responses, reflecting the most common API outputs encountered in real-world scenarios. Finally, we assume that the graph $\mathbf{G}_D = (\mathbf{X}_D, \mathbf{A}_D)$ is drawn from the same distribution as the graph $\mathbf{G}_V$, which is used for training the target model $f_v$ (\reb{which is a standard assumption in the literature \cite{shen2021model, Podhajski_2024, wu2021model,TZJRR16,JCBKP20}}). In practice, we consider $\mathbf{G}_D$ and $\mathbf{G}_V$ to come from the same distribution if they are sampled from the same dataset.

\subsection{Obtaining the Encoder}

In the initial stage of our method, we deliberately avoid making any queries to the victim model. Instead, we focus on obtaining the encoder part of the model locally, allowing us to thoroughly analyze and understand the data without prematurely utilizing resources on victim model queries. The encoder obtained in this step serves a twofold purpose:
1) we reuse the encoder as a component of the surrogate model, and
2) it allows us to comprehend the data and select the best possible queries in the subsequent step of our method.

In the inductive setting, we start from the observation that various studies on SSL training, including DGI \cite{velickovic2018deep}, BGRL \cite{thakoor2022large}, and LaGraph \cite{pmlr-v162-xie22e}, report that a randomly initialized encoder (particularly using GCN architecture) often yields results comparable to those of an SSL-trained encoder. 
Our observations align with these findings. We hypothesize that a randomly initialized encoder can often achieve similar performance to an SSL-trained encoder. The results presented in the experimental section positively verify this hypothesis.
This leads to a novel paradigm for the inductive setting in which a randomly initialized model can achieve results comparable to those of stolen models without needing direct queries to the victim model. On top of this, using our approach, even without significant computational resources, one can obtain a good encoder for large graph datasets. 

In the transductive setting, the randomly initialized GCN was observed to have lower performance \cite{velickovic2018deep}. 
Therefore, in this setting, we utilize an SSL approach. We note that in the transductive setting, the graph used for training is typically small, enabling us to train 
the encoder locally via SSL without extensive resource consumption. This is advantageous because it ensures that our initial model training is efficient and cost-effective. Additionally, this procedure ensures that the adversary utilizes all of the data points that it has access to, regardless of the strict limit of queries to the victim. 

Our approach is tailored to be compatible with any SSL framework that produces an encoder capable of generating meaningful representations. This flexibility allows us to integrate our approach with a variety of existing SSL techniques, ensuring broad applicability and effectiveness. In particular, in our numerical experiments, as the SSL framework we employ LaGraph \cite{pmlr-v162-xie22e}. In the inductive setting, as our encoder, we applied the GCN architecture following the approach proposed by \citet{pmlr-v162-xie22e}. In the transductive setting, we use the SSL approach, which is effective for a range of backbone GNNs. For our experiments, we utilize both GCN, GIN \cite{XHLJ19}, and SAGE architectures.

\subsection{Query Selection}

In the final step, we aim to select a data subset that maximizes information extraction from the victim model within the query limit. Intuitively, our goal is to select the most diverse subset possible, ensuring comprehensive coverage of the dataset accessible to the adversary. This strategy of selecting representative points to cover the input space is conceptually analogous to diversity sampling methods in the field of active learning \cite{settles_active_2009, 53258}. This selected subset is then used to train an MLP head on top of the encoder, enabling the construction of a high-quality surrogate model.

Graph data $\mathbf{G}_D$ available to the adversary consists of a set of attributes $\mathbf{X}_D \in \mathcal{R}^{n*d}$ and a graph structure $\mathbf{A}_D \in \{0, 1\}^{n \times n}$, where $n$ is the number of nodes and $d$ is the number of features. To cover the dataset well, we want to map $\mathbf{X}_D$ and $\mathbf{A}_D$: $f(\mathbf{X}_D, \mathbf{A}_D)$ into a space that will contain information about both parts. For this purpose, we can use an encoder $f$ that maps the attributes and structure of the graph into the $\mathbf{H} \in \mathcal{R}^{n*b}$ embedding space, where $b$ is the embedding size. Note that for the inductive setting, we use a randomly initialized encoder, and for the transductive setting, we use an SSL-trained model. The training set for the SSL-encoder comprises all data available to the adversary, denoted as the set of nodes $\{v_1, ..., v_n\}$.

To choose a subset of nodes that covers the embedding space $\mathbf{H}$, we use the K-means algorithm \cite{Lloyd1982LeastSQ}. We partition the embeddings $\{h_1, ..., h_n\} \subset \mathbf{H}$ of nodes $\{v_1, ..., v_n\}$ into $q_n$ clusters, each containing at least one node. The number of clusters $q_n$ is set equal to the victim model's query limit, corresponding to the number of queries we are allowed to send to the target. Next, we select one node from each cluster (specifically, the one whose embedding is closest to the cluster centroid), forming the set of nodes $\{v'_1, ..., v'_{q_n}\}$ for our query graph $\mathbf{G}_Q$. Finally, we query the victim model $f_v$ and obtain predictions $\{y_1, ..., y_{q_n}\}$.

In the last step, we train an MLP component $g$ to predict a class $y$ based on the representation $h$ using Cross Entropy (CE) loss. The training set of $g$ consists of the chosen embeddings $\{h'_1, ..., h'_{q_n}\} = \{f(v'_1), ...,f(v'_{q_n})\}$ and labels returned by the victim $\{y_1, ..., y_{q_n}\}$.
The final surrogate model $f_s$ consists of the encoder $f$ and a head $g$ which makes predictions based on a graph structure $\mathbf{A}$ and attributes $\mathbf{X}$: $\hat{y} = f_s(\mathbf{X},\mathbf{A}) =  g(f(\mathbf{X},\mathbf{A}))$. 

\section{Empirical Evaluation}
\label{sec:empirical_evaluation}
We evaluate our proposed approach across eight benchmark datasets to demonstrate its effectiveness and to highlight the vulnerability of GNNs to model-stealing attacks in both transductive and inductive settings. The experimental setup is described in detail in
\ifappendixincluded
Section \ref{exp_setup} in the Appendix\fi.

\begin{table*}[htbp]
\centering
\small
\caption{Inductive setting (target: SAGE, surrogate: GCN (same for all methods), $q_n = 100$). Accuracy (Acc.) and Fidelity (Fid.) are reported as mean ± std.\ dev.\ in percentage over 3 runs. Methods marked with * assume weaker threat model (access to victim embeddings).}
\label{tab:acc_fidelity_random_ind}
\setlength{\tabcolsep}{1pt} 
\begin{tabular}{lcccccccccc}
\toprule
\textbf{Method} &
\multicolumn{2}{c}{\textbf{Reddit}} &
\multicolumn{2}{c}{\textbf{CS}} &
\multicolumn{2}{c}{\textbf{Physics}} &
\multicolumn{2}{c}{\textbf{Photo}} &
\multicolumn{2}{c}{\textbf{WikiCS}} \\
\cmidrule(lr){2-3}
\cmidrule(lr){4-5}
\cmidrule(lr){6-7}
\cmidrule(lr){8-9}
\cmidrule(lr){10-11}
 & \textbf{Acc.} & \textbf{Fid.} & \textbf{Acc.} & \textbf{Fid.} & \textbf{Acc.} & \textbf{Fid.} & \textbf{Acc.} & \textbf{Fid.} & \textbf{Acc.} & \textbf{Fid.} \\
\midrule
Target accuracy & 94.8 &  & 93.9 &  & 96.0 &  & 93.0 &  & 72.5 &  \\
\cmidrule(lr){2-11}
\textit{E2E} & 
47.0{$\pm$4.5} & 47.0{$\pm$4.4} & 
73.6{$\pm$3.9} & 74.5{$\pm$3.9} & 
89.9{$\pm$1.1} & 91.1{$\pm$1.3} & 
81.2{$\pm$0.8} & 83.3{$\pm$1.7} & 
61.6{$\pm$1.3} & 71.0{$\pm$2.5} \\

\textit{R-init + Random} & 
76.9{$\pm$4.0} & 77.0{$\pm$3.9} & 
74.2{$\pm$2.2} & 74.5{$\pm$2.3} & 
86.0{$\pm$1.5} & 87.8{$\pm$1.5} & 
85.0{$\pm$1.7} & 87.8{$\pm$1.8} & 
63.0{$\pm$2.0} & 71.4{$\pm$2.4} \\
\cmidrule(lr){2-11}

\citet{shen2021model} & 
77.2*{$\pm$5.1} & 77.0*{$\pm$4.5} & 
77.7*{$\pm$0.8} & 78.7*{$\pm$0.7} & 
90.6*{$\pm$0.5} & 91.6*{$\pm$0.6} & 
84.4*{$\pm$0.8} & 86.3*{$\pm$0.8} & 
64.9*{$\pm$1.0} & \textbf{81.0}*{$\pm$0.9} \\

\citet{Podhajski_2024} & 
79.9*{$\pm$4.1} & 79.5*{$\pm$4.4} & 
78.0*{$\pm$0.5} & 78.1*{$\pm$0.4} & 
89.9*{$\pm$0.2} & 89.1*{$\pm$0.3} & 
84.0*{$\pm$1.0} & 84.2*{$\pm$1.2} & 
64.0*{$\pm$1.1} & 70.0*{$\pm$0.5} \\

\citet{datafree} & 
13.6{$\pm$4.1} & 19.5{$\pm$3.2} & 
24.8{$\pm$2.8} & 27.1{$\pm$3.9} & 
55.5{$\pm$5.0} & 54.9{$\pm$4.6} & 
24.9{$\pm$2.8} & 24.9{$\pm$3.2} & 
38.6{$\pm$2.1} & 40.8{$\pm$2.0} \\

\textbf{R-init + Select (ours)} & 
\textbf{82.5}{$\pm$1.2} & \textbf{82.7}{$\pm$1.2} & 
\textbf{78.4}{$\pm$2.1} & \textbf{79.2}{$\pm$2.2} & 
\textbf{91.2}{$\pm$0.4} & \textbf{92.7}{$\pm$0.5} & 
\textbf{86.8}{$\pm$1.0} & \textbf{89.8}{$\pm$0.9} & 
\textbf{65.5}{$\pm$1.8} & 73.6{$\pm$1.9} \\
\bottomrule
\end{tabular}

\end{table*}

\subsection{Stealing with a Random Encoder and Self-Supervised Learning}
\label{sec:empirical_encoder}
We first consider the inductive setting and investigate the efficacy of using a randomly initialized encoder. Specifically, we compare two approaches:
\begin{itemize}
    \item training a multi-layer perceptron on top of a randomly initialized and frozen encoder (\textbf{R-init}), and
    \item training the entire GNN architecture, including both the encoder and the head, in an end-to-end manner (\textbf{E2E}).
\end{itemize}

The results of this comparison, as presented in
\ifappendixincluded
Tables \ref{tab:acc_fidelity_random_ind} and \ref{tab:acc_fidelity_random_ind_gat} (Appendix)\fi, include the accuracy and fidelity of both approaches, reported along with their standard deviations. These evaluations were conducted using randomly selected queries, without leveraging any query selection algorithm. The results reveal that the performance of the frozen encoder with an MLP is comparable to that of the \textit{E2E} approach across most datasets, as well as when using both GAT and SAGE target models. Additionally, 
\ifappendixincluded
Figures \ref{fig:grp1} and \ref{fig:grp2} (Appendix)\fi ~present charts that confirm these results hold across different query limits. Importantly, the R-init method requires significantly fewer computational resources, as only the MLP layer needs training, whereas the \textit{E2E} approach involves updating the entire network. 
We used T-SNE to visualize embeddings from the Physics dataset, as shown in  
\ifappendixincluded
Figure \ref{fig:tsne_combined} (Appendix)\fi. 
The 2D plot compares embeddings from a random encoder and a trained encoder. Clusters corresponding to node classes are clearly separated, with the random encoder producing well-defined clusters closely resembling those of the trained encoder. 

In addition to the inductive setting, we also explore the transductive setting by evaluating the impact of self-supervised learning on encoder performance. Beyond the \textit{E2E} approach, we consider an SSL-trained encoder paired with an MLP (\textbf{SSL}). The results, summarized in 
\ifappendixincluded
Tables \ref{tab:combined_results} and \ref{tab:combined_results_gat} (\reb{in the Appendix)}\fi, demonstrate a significant improvement in accuracy and fidelity when using SSL compared to relying solely on the query data set. This improvement is particularly notable in datasets such as Cora and Citeseer and is observed consistently across different target models (GCN and GAT) as well as surrogate architectures (GIN, GCN, and SAGE). 
\ifappendixincluded
Figure \ref{fig:grp3} (\reb{in the Appendix)}\fi  ~confirms this result across different query limits, showing that the performance improvement from SSL becomes more pronounced as the query limit decreases. In addition, the findings suggest that incorporating all available adversarial data in local self-supervised training yields substantial performance gains across all benchmarks. This underscores the utility of SSL in extracting high-quality representations, even in extreme scenarios, enhancing the overall robustness and effectiveness of the model-stealing process.

\begin{table*}[!t]
\centering
\small
\caption{Transductive setting (target: GCN, surrogates: GIN, SAGE, GCN, $q_n = 10$). Accuracy (Acc.) and Fidelity (Fid.) reported as mean ± std.\ dev.\ in percentage over 3 runs.}
\label{tab:combined_results}
\setlength{\tabcolsep}{3pt} 
\begin{tabular}{ll 
                cc 
                cc 
                cc}
\toprule
\multicolumn{2}{c}{} & \multicolumn{2}{c}{\textbf{Cora}} & \multicolumn{2}{c}{\textbf{Citeseer}} & \multicolumn{2}{c}{\textbf{Pubmed}}\\
\cmidrule(lr){3-4} \cmidrule(lr){5-6} \cmidrule(lr){7-8}
\textbf{Surrogate} & \textbf{Method} & \textbf{Acc.} & \textbf{Fid.} & \textbf{Acc.} & \textbf{Fid.} & \textbf{Acc.} & \textbf{Fid.} \\
\midrule
   & Target accuracy & 83.3 & & 72.1 & & 80.0 & \\
\midrule
\multirow{4}{*}{\textbf{GIN}} 
   & \textit{E2E} \cite{wu2021model}       & 39.1{$\pm$9.7} & 38.8{$\pm$8.0} & 37.8{$\pm$3.1} & 40.1{$\pm$5.5} & 57.1{$\pm$3.1} & 66.0{$\pm$3.0} \\
   & \citet{datafree}                     & 17.7{$\pm$4.4} & 21.0{$\pm$5.1} & 23.3{$\pm$2.8} & 26.0{$\pm$4.0} & 35.1{$\pm$2.2} & 36.6{$\pm$4.0} \\
   & \textit{SSL + Random}                & 52.7{$\pm$4.7} & 51.5{$\pm$6.1} & 63.3{$\pm$2.1} & 56.5{$\pm$3.5} & 69.1{$\pm$4.3} & 72.5{$\pm$4.0} \\
   & \textbf{SSL + Select (ours)}         & {57.7}{$\pm$2.9} & {57.8}{$\pm$2.3} & {65.6}{$\pm$1.0} & {71.5}{$\pm$1.2} & {69.9}{$\pm$3.1} & {76.0}{$\pm$3.0} \\
\midrule
\multirow{4}{*}{\textbf{SAGE}} 
   & \textit{E2E} \cite{wu2021model}       & 46.2{$\pm$2.2} & 35.8{$\pm$5.1} & 27.8{$\pm$4.9} & 25.9{$\pm$7.6} & {64.1}{$\pm$0.9} & 63.1{$\pm$6.1} \\
   & \citet{datafree}                     & 11.5{$\pm$3.0} & 10.9{$\pm$3.1} & 13.9{$\pm$2.9} & 12.3{$\pm$4.6} & 36.9{$\pm$2.7} & 41.0{$\pm$3.0} \\
   & \textit{SSL + Random}                & 40.8{$\pm$7.5} & 42.9{$\pm$7.4} & 44.5{$\pm$7.1} & {50.5}{$\pm$6.8} & 62.8{$\pm$3.5} & 65.4{$\pm$3.0} \\
   & \textbf{SSL + Select (ours)}         & {46.8}{$\pm$5.6} & {46.5}{$\pm$4.5} & {45.7}{$\pm$4.5} & 47.7{$\pm$7.1} & {62.8}{$\pm$1.6} & {69.4}{$\pm$1.2} \\
\midrule
\multirow{4}{*}{\textbf{GCN}} 
   & \textit{E2E} \cite{wu2021model}       & 47.5{$\pm$3.7} & 45.7{$\pm$1.0} & 37.2{$\pm$6.1} & 41.1{$\pm$7.5} & 61.0{$\pm$4.9} & 67.5{$\pm$5.0} \\
   & \citet{datafree}                     & 18.1{$\pm$2.7} & 21.1{$\pm$3.9} & 22.1{$\pm$3.3} & 23.1{$\pm$3.8} & 33.2{$\pm$2.9} & 33.4{$\pm$3.0} \\
   & \textit{SSL + Random}                & 56.1{$\pm$2.7} & 56.8{$\pm$3.0} & 51.3{$\pm$5.1} & 57.6{$\pm$5.5} & {66.1}{$\pm$7.3} & 72.7{$\pm$9.0} \\
   & \textbf{SSL + Select (ours)}         & \textbf{69.9}{$\pm$1.2} & \textbf{72.5}{$\pm$1.3} & \textbf{66.3}{$\pm$1.9} & \textbf{72.4}{$\pm$2.3} & \textbf{67.0}{$\pm$6.0} & \textbf{80.1}{$\pm$4.7} \\
\bottomrule
\end{tabular}

\end{table*}

\subsection{Strategic Query Selection}
\label{sec:empirical_queries}
Finally, we evaluate the ultimate step of the method proposed in both the transductive setting
\ifappendixincluded
(see Table \ref{tab:combined_results} and \reb{Appendix Table} \ref{tab:combined_results_gat}) and the inductive setting (see Table \ref{tab:acc_fidelity_random_ind} and \reb{Appendix Table} \ref{tab:acc_fidelity_random_ind_gat})\fi. For comparison, we use:
\begin{itemize}
\item a randomly chosen set of queries (\textbf{Random}), and
\item a set of queries selected using our method (\textbf{Select}).
\end{itemize}
The results show that selecting queries with K-means based on the encoder embeddings results in higher accuracy and fidelity in both setups across all of the datasets and target models. The results also indicate that in the transductive setting, our method provides a selection of more diverse queries based on the SSL-trained embeddings. For the inductive setting, it is further confirmed that the representations produced by the random encoder are meaningful and possible to interpret. By comparing the accuracy of our surrogate models with the performance of the victim models, we observe that the surrogate models achieve comparable results with significantly fewer labeled data. Additionally, 
\ifappendixincluded
Figures \ref{fig:grp1}, \ref{fig:grp2} and \ref{fig:grp3} \reb{in the Appendix}\fi ~illustrate how performance improves with different query limits, further demonstrating that as the query limit decreases, the effectiveness of our method increases. To justify the use of our approach in the \textit{Select} phase, we compare K-means with several representative selection strategies, including farthest-first, K-center greedy, entropy sampling, coreset herding, and margin sampling~\cite{settles_active_2009, scheffer2001active, welling2009herding, gonzalez1985clustering}. As shown in 
\ifappendixincluded
Table~\ref{tab:active_learning_results} in the Appendix\fi, while methods such as coreset herding offer improvements over random selection, K-means consistently delivers the highest accuracy and fidelity across all datasets.

To quantitatively assess the properties of the selected query set, we perform an experiment measuring the fraction of class coverage per query. Based on the average of 100 runs on the CS dataset
\ifappendixincluded
(Figure \ref{fig:random_vs_select} \reb{in the Appendix})\fi, we observe that our selection method covers a greater number of classes than random sampling under small budgets, with both approaches converging to full class coverage as the query limit increases. Additionally,
\ifappendixincluded
Figure \ref{fig:selected} \reb{in the Appendix}\fi
~presents a T-SNE projection illustrating the distribution of nodes selected by our query strategy.

We also conduct McNemar’s test \cite{McNemar1947}, as presented in
\ifappendixincluded
Tables \ref{tab:pvalues} and \ref{tab:pvalues_trans} \reb{in the Appendix}\fi, to compare the stolen model with the original model by evaluating their classification errors on the same dataset. The null hypothesis assumes no significant difference in the classification error rates between the stolen and original models. The results demonstrate that our method produces a stolen model with a higher degree of similarity to the victim model.

\subsection{Comparison with Existing Methods}

We thoroughly compare our method with existing approaches for stealing inductive and transductive GNNs.

\textbf{Inductive setting.} We compare our method against all existing approaches, \textit{i.e.}, \citet{shen2021model} and \citet{Podhajski_2024}. We note that these two previous approaches do not take into account scenarios where the adversary's dataset exceeds the query limit. Thus, to explore this scenario, we replicate the experiments described in \citet{shen2021model} and \citet{Podhajski_2024} with randomly selected queries from the entire available dataset. 
Additionally, we compare our performance with \citet{datafree}, which demonstrates that an adversary can successfully steal a model without access to any training data, though this requires a substantial number of queries: 100 queries for graphs of size 250, totaling 25,000 query nodes. We show that our method performs much better in a scenario with very restricted model access (we evaluate our approach on 10, 25, 50, 100, and 500 nodes).
We present our results in
\ifappendixincluded
Table \ref{tab:acc_fidelity_random_ind} and \reb{Appendix Table} \ref{tab:acc_fidelity_random_ind_gat}\fi, where we compare all methods using the same surrogate model architecture and different target architectures: SAGE and GAT, with a query limit $q_n$ of 100. We also compare the performance of all methods across different query limits $q_n$ in
    \ifappendixincluded
    Figures \ref{fig:grp1} and \ref{fig:grp2} (Appendix)\fi. Furthermore, \ifappendixincluded
    in Table \ref{tab:gin_sage_gat} (Appendix)\fi, we compare the performance using the architectures originally employed in \citet{shen2021model} and \citet{Podhajski_2024}. Our method, which does not rely on query embeddings from the victim model, consistently outperforms the previous methods in both accuracy and fidelity. This superiority is evident across various baseline surrogate models, including GIN, GAT, and SAGE. Additionally, by utilizing the randomly initialized encoder, we significantly improved the execution times as seen \ifappendixincluded
    in Table \ref{tab:gin_sage_gat} (Appendix)\fi. Moreover, even when the same architecture is used, our approach yields significantly better results, demonstrating its effectiveness.

\textbf{Transductive setting.} For the transductive setting, there is no previous work in the literature to which we can directly compare. However, we believe that in the work by \citet{wu2021model} we can find a setting to which a comparison of our \textit{E2E} is reasonable. In particular, unlike our work, \citet{wu2021model} focuses on the scenarios where the adversary lacks some types of data, e.g., the graph structure or node attributes. In contrast, our focus is on scenarios where the adversary is not limited in terms of data type but in terms of the number of queries that can be made. To create a relevant comparison, we modified the approach by \citet{wu2021model} by training their model using randomly selected queries under the assumption that it has access to the entire data available to the adversary (with no restrictions on data types). 
Similarly, we compare our results to \citet{datafree}, which generates graph queries in a restricted scenario, where the query limit is reduced to just 5, 10, 20, or 50 nodes.
The results presented in
\ifappendixincluded Tables \ref{tab:combined_results} and \ref{tab:combined_results_gat}  (\reb{in the Appendix})\fi ~show a significant improvement in both accuracy and fidelity when our method is applied. Specifically, when the surrogate encoder is trained with all available data and the query selection is optimized using the K-means clustering algorithm, the performance of the stolen model is greatly enhanced. This demonstrates the advantage of our approach in transductive settings, where strategic data utilization and query selection can significantly boost the attack's effectiveness. We further validate the effectiveness of our method across different surrogate architectures (GIN, SAGE, and GCN) and observe that our method consistently outperforms all baselines regardless of the surrogate used. Notably, the GCN surrogate yields the highest performance overall, surpassing all other methods and architectural configurations. Additionally,
\ifappendixincluded Figure \ref{fig:grp3}\fi ~shows how the performance improves with different query limits, further emphasizing the impact of our method as the query limit decreases.

\section{Defense}
In this paper, we demonstrate that our method can successfully perform a model extraction even under very restrictive conditions, where the victim model returns only the predicted class and there are a limited number of queries that the adversary can make.  Although numerous defenses against model extraction have been proposed in the literature \cite{TZJRR16,dziedzic2022increasing, pmlr-v162-dziedzic22a, 10080996}, they typically assume a less restrictive setup. For example, if the victim model returns the embeddings, a possible defense is to inject a random Gaussian noise into the node embeddings (the setup studied by \citet{Podhajski_2024, shen2021model, dubiński2023bucksbucketsb4bactive}). Similarly, when returning class probabilities, some methods \cite{10080996, orekondy20prediction} perturb the output distribution. In contrast, in our hard-label setting, a possible defense \cite{TZJRR16} is based on changing the class prediction (for example, with some probability $p$). However, such defenses come with the cost of reducing the accuracy of the model.

We evaluate all stealing methods under a defense that flips predictions with a probability of $p = 10\%$ in both inductive and transductive settings 
\ifappendixincluded (Tables~\ref{tab:defense_ind} and~\ref{tab:defense_trans} \reb{in the Appendix})\fi. Our proposed method consistently achieves the highest performance, even in the presence of defense measures.
These results highlight the difficulty of defending against our extraction attack, even under strong defense mechanisms that come at the cost of the utility of the model.

\section{Conclusions}
In this paper, we studied and challenged traditional approaches to stealing GNN models within the frameworks of both inductive and transductive settings.

We examined existing GNN model theft methods in the \textbf{inductive setting}, which typically involve the extraction of an encoder and training of an MLP using class labels. Our analysis reveals that models initialized randomly can yield results comparable to those trained with additional responses (e.g., embeddings) from the victim. This approach demonstrates the potential for effective model extraction, even with low computational resources and restricted victim access.

We extend our analysis to the \textbf{transductive setting}, showing that when the adversary’s data surpasses the query limit, the excess can be used via SSL to boost the stolen model’s performance. Our results demonstrate that even with limited queries, leveraging extra data significantly improves the stolen model’s accuracy and fidelity.

Additionally, in \textbf{both settings}, we studied how the adversary can take advantage of having more data than the query limit. We demonstrated that by strategically selecting optimal nodes for querying, it is possible to significantly enhance the accuracy and fidelity of the stolen model.

\par
We believe that our research offers novel insights into GNN model-stealing techniques across both inductive and transductive frameworks. The results obtained not only contribute to a deeper understanding of these attacks but also highlight the urgent need for improved security measures against such adversarial strategies.

\bibliography{aaai2026}


\ifappendixincluded

\appendix
\clearpage
\section{Experimental Setup}
\label{exp_setup}
\par
\textbf{Inductive Setting.} For the inductive setting, we conduct experiments on Physics, CS~\cite{SMBG18}, Photos~\cite{MTSH15}, Reddit~\cite{graphsaint-iclr20}, and WikiCS~\cite{mernyei2020wiki} datasets. These benchmarks are standard for inductive evaluation and were chosen to test our attack's scalability across a variety of graph sizes and domains. Their diversity, from large social networks like Reddit to co-purchase graphs like Photos, ensures a robust test of generalizability. Following a standard practice to ensure robust evaluation, the datasets are partitioned into $40.0$\% for training, $10.0$\% for validation, and $50.0$\% for testing. Specifically, for Reddit, we employ the public split provided by PyTorch Geometric, maintaining consistency with previous studies.
In this setting, we evaluate our attack on two target models: the first one is a $5$-layer SAGE with a hidden size of $512$, ReLU activation, and a dropout rate of $0.5$. Similarly, we consider a $5$-layer GAT model with a hidden size of $128$, $4$ heads, ReLU activation, and a dropout rate of $0.5$ --- resulting in the same embedding size as the other models, which is $512$. The surrogate model is a $5$-layer GCN with a matching hidden size of $512$, batch normalization, and PReLU activation. The architectures were chosen for their empirical performance on these tasks. However, the proposed model stealing methodology does not rely on any specific architectural congruence between the surrogate and the victim. Network details or other hyperparameters are not prerequisites for the attack, as it exclusively utilizes the hard label predictions returned by the victim API. Therefore, any similarities between an effective surrogate and a potential victim are not a dependency of the stealing process itself.

\textbf{Transductive Setting.} For the transductive experiments, we utilize the widely-recognized Citeseer~\cite{GBL98}, Pubmed~\cite{SNBGGE08}, and Cora~\cite{mccallum2000automating} datasets. These three are canonical citation networks that form the standard testbed for transductive GNNs. Their inclusion allows for direct and fair comparison against prior work. The datasets are divided according to the public split protocol provided by \cite{Yang2016RevisitingSL}, ensuring consistency with prior work and enabling a fair comparison. In this setting, the target models consist of two architectures. The first is a $2$-layer GCN with a hidden size of 256, a dropout rate of $0.5$, and ReLU activation functions. The second target model is a $2$-layer GAT with a hidden size of $128$, $4$ attention heads, a dropout rate of $0.5$, and ReLU activation functions. For the surrogate models used by the adversary, we employ a $2$-layer GCN with a dropout rate of 0.5 and ReLU activation, a $2$-layer Graph Isomorphism Network (GIN) with ReLU activation, and a $2$-layer SAGE with a dropout rate of 0.5 and ReLU activation. In the SSL training, we follow LaGraph's hyperparameter setup with a mask ratio of $0.05$, a learning rate of $0.001$, $30$ epochs, an Adam optimizer \cite{KingBa15}, and a one-layer MLP decoder.
In the final MLP training, we use a one-layer network, and we employ an Adam optimizer with a learning rate of $0.01$ and $100$ epochs. To ensure a fair and reproducible setup, the hyperparameters for the victim models and the SSL encoder were adopted directly from their respective original papers. The final MLP parameters mentioned above were selected via a grid search over learning rates $\{0.01, 0.001\}$ and epochs $\{50, 100, 200, 500\}$, with the criterion being to maximize the surrogate model's performance on the queried data.

\textbf{Evaluation metrics.} We use two metrics from the \cite{JCBKP20} taxonomy to evaluate our attack: accuracy and fidelity. \textit{Accuracy}, the ratio of correct to total predictions, measures the theft adversary’s performance. \textit{Fidelity}, which reflects the agreement between the predictions of the surrogate and target models, assesses the reconnaissance adversary. All experiments are conducted three times, and we calculate the standard deviation to ensure reliability; the specific seeds are set in our publicly available source code

Experiments were conducted on Ubuntu 22 with Python 3.10 and PyTorch 1.13, using a single commercial-grade AMD EPYC 7742 CPU. This showcases the practicality of executing our approach with modest computational resources, emphasizing the threat of model-stealing attacks.

\section{Additional Experimental Results}
\subsection{Encoder Embeddings Visualisation}
\begin{figure}[h!]
    \centering

    \subfloat[\textbf{Inductive (Physics)}\\Trained on victim embeddings]{
        \includegraphics[width=0.49\linewidth]{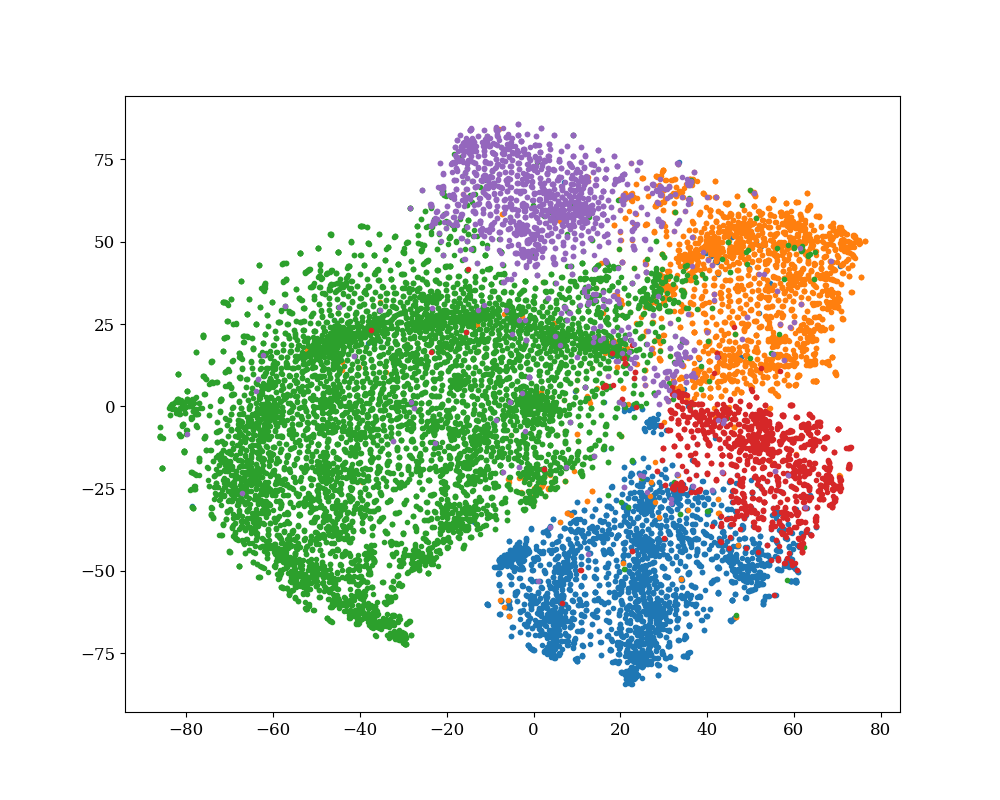}
    }
    \subfloat[\textbf{Inductive (Physics)}\\Randomly initialized]{
        \includegraphics[width=0.49\linewidth]{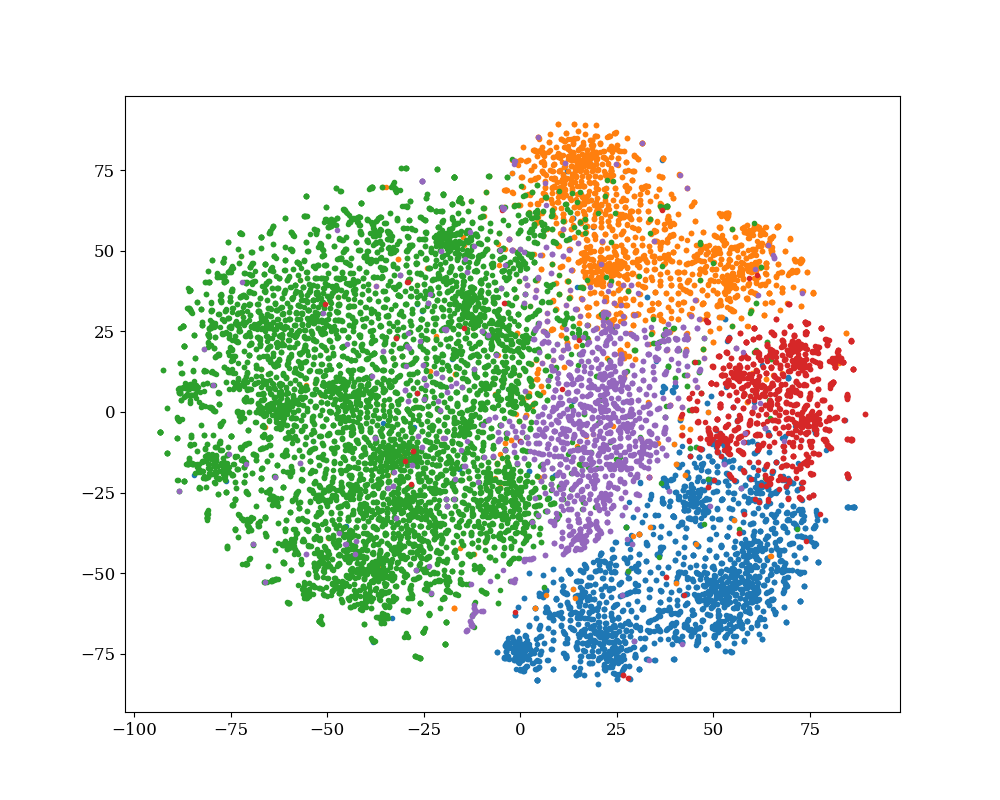}
    }

    \par\medskip

    \subfloat[\textbf{Transductive (Citeseer)}\\Trained with SSL]{
        \includegraphics[width=0.49\linewidth]{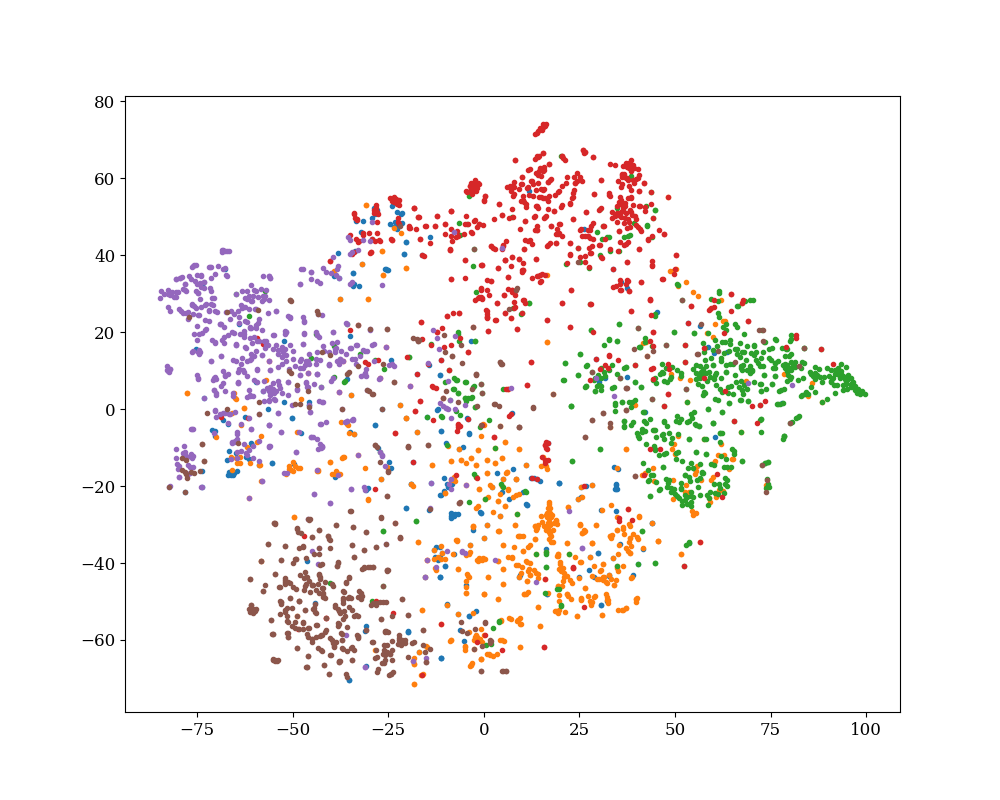}
    }
    \subfloat[\textbf{Transductive (Citeseer)}\\Randomly initialized]{
        \includegraphics[width=0.49\linewidth]{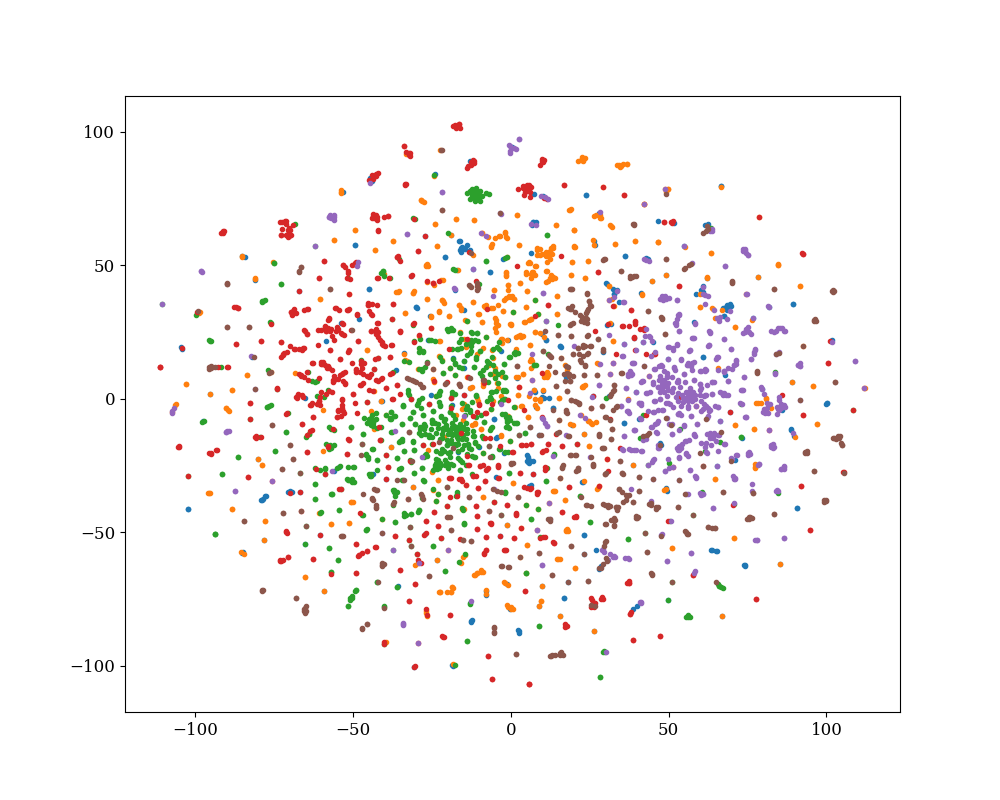}
    }

    \caption{T-SNE projections of embeddings on the Citeseer dataset (transductive setting) and the Physics dataset (inductive setting).}
    \label{fig:tsne_combined}
\end{figure}

\onecolumn

\subsection{Full Evaluation Results}

\begin{table*}[!h]
\centering
\small
\caption{Inductive setting (target: GAT, surrogate: GCN (same for all methods), $q_n = 100$). Accuracy (Acc.) and Fidelity (Fid.) are reported as mean ± std.\ dev.\ in percentage over 3 runs. Methods marked with * assume a weaker threat model (access to victim embeddings).}
\label{tab:acc_fidelity_random_ind_gat}
\setlength{\tabcolsep}{1pt} 
\begin{tabular}{lcccccccccc}
\toprule
\textbf{Method} &
\multicolumn{2}{c}{\textbf{Reddit}} &
\multicolumn{2}{c}{\textbf{CS}} &
\multicolumn{2}{c}{\textbf{Physics}} &
\multicolumn{2}{c}{\textbf{Photo}} &
\multicolumn{2}{c}{\textbf{WikiCS}} \\
\cmidrule(lr){2-3}
\cmidrule(lr){4-5}
\cmidrule(lr){6-7}
\cmidrule(lr){8-9}
\cmidrule(lr){10-11}
 & \textbf{Acc.} & \textbf{Fid.} & \textbf{Acc.} & \textbf{Fid.} & \textbf{Acc.} & \textbf{Fid.} & \textbf{Acc.} & \textbf{Fid.} & \textbf{Acc.} & \textbf{Fid.} \\
\midrule
Target accuracy & 93.8 &  & 92.0 &  & 94.8 &  & 89.9 &  & 75.8 &  \\
\cmidrule(lr){2-11}
\textit{E2E} &
42.6{$\pm$6.8} & 43.1{$\pm$5.9} &
79.7{$\pm$1.0} & 80.7{$\pm$1.1} &
88.1{$\pm$0.3} & 89.2{$\pm$0.3} &
74.4{$\pm$0.6} & 76.9{$\pm$0.6} &
59.7{$\pm$4.8} & 67.3{$\pm$4.1} \\

\textit{R-init + Random} &
78.5{$\pm$1.6} & 78.5{$\pm$1.8} &
77.1{$\pm$0.8} & 78.1{$\pm$0.7} &
86.6{$\pm$3.1} & 87.5{$\pm$3.2} &
85.3{$\pm$1.1} & 89.0{$\pm$1.2} &
60.2{$\pm$1.0} & 70.8{$\pm$1.2} \\
\cmidrule(lr){2-11}

\citet{shen2021model} &
74.4*{$\pm$3.0} & 74.6*{$\pm$3.1} &
77.4*{$\pm$2.2} & 77.2*{$\pm$2.2} &
90.5*{$\pm$0.8} & 91.5*{$\pm$0.8} &
84.2*{$\pm$0.4} & 86.0*{$\pm$0.4} &
56.7*{$\pm$1.0} & \textbf{81.6}*{$\pm$2.1} \\

\citet{Podhajski_2024} &
76.0*{$\pm$4.0} & 74.0*{$\pm$2.8} &
78.4*{$\pm$2.9} & 79.6*{$\pm$2.6} &
88.6*{$\pm$1.5} & 79.9*{$\pm$1.8} &
84.3*{$\pm$1.3} & 82.5*{$\pm$2.1} &
56.6*{$\pm$1.2} & 73.1*{$\pm$3.3} \\

\citet{datafree} &
18.4{$\pm$3.1} & 19.8{$\pm$4.0} &
23.4{$\pm$17.4} & 12.8{$\pm$6.8} &
43.1{$\pm$24.7} & 37.4{$\pm$18.5} &
25.0{$\pm$1.1} & 24.9{$\pm$1.2} &
23.0{$\pm$1.4} & 22.0{$\pm$1.8} \\

\textbf{R-init + Select (ours)} &
\textbf{82.2}{$\pm$0.6} & \textbf{82.3}{$\pm$0.6} &
\textbf{80.1}{$\pm$0.8} & \textbf{81.1}{$\pm$0.8} &
\textbf{91.4}{$\pm$0.6} & \textbf{92.3}{$\pm$0.5} &
\textbf{85.7}{$\pm$0.5} & \textbf{89.9}{$\pm$0.5} &
\textbf{63.6}{$\pm$0.7} & 74.2{$\pm$0.9} \\
\bottomrule
\end{tabular}

\end{table*}

\begin{table*}[!h]
\centering
\small
\caption{Transductive setting (target: GAT, surrogates: GIN, SAGE, GCN, $q_n = 10$). Accuracy (Acc.) and Fidelity (Fid.) are reported as mean ± std.\ dev.\ in percentage over 3 runs.}
\label{tab:combined_results_gat}
\setlength{\tabcolsep}{3pt} 
\begin{tabular}{ll 
                cc 
                cc 
                cc}
\toprule
\multicolumn{2}{c}{} & \multicolumn{2}{c}{\textbf{Cora}} & \multicolumn{2}{c}{\textbf{Citeseer}} & \multicolumn{2}{c}{\textbf{Pubmed}}\\
\cmidrule(lr){3-4} \cmidrule(lr){5-6} \cmidrule(lr){7-8}
\textbf{Surrogate} & \textbf{Method} & \textbf{Acc.} & \textbf{Fid.} & \textbf{Acc.} & \textbf{Fid.} & \textbf{Acc.} & \textbf{Fid.} \\
\midrule
   & Target accuracy & 80.6 &  & 70.3 &  & 77.7 & \\
\midrule
\multirow{4}{*}{\textbf{GIN}} 
   & \textit{E2E} \cite{wu2021model}       & 42.9{$\pm$1.5} & 36.1{$\pm$6.3} & 39.8{$\pm$4.6} & 33.3{$\pm$7.3} & 53.3{$\pm$7.6} & 49.2{$\pm$7.0} \\
   & \citet{datafree}                     & 12.1{$\pm$1.5} & 12.0{$\pm$2.4} & 20.8{$\pm$4.6} & 19.2{$\pm$4.7} & 40.5{$\pm$2.3} & 40.9{$\pm$3.5} \\
   & \textit{SSL + Random}                & 46.9{$\pm$10.7} & 47.0{$\pm$10.1} & 44.8{$\pm$4.1} & 44.9{$\pm$4.6} & 61.9{$\pm$4.8} & 64.9{$\pm$6.6} \\
   & \textbf{SSL + Select (ours)}         & {51.8}{$\pm$6.2} & {52.0}{$\pm$5.6} & {60.3}{$\pm$2.4} & {64.7}{$\pm$3.8} & {64.0}{$\pm$2.7} & {69.1}{$\pm$3.7} \\
\midrule
\multirow{4}{*}{\textbf{SAGE}} 
   & \textit{E2E} \cite{wu2021model}       & 43.1{$\pm$11.8} & 32.8{$\pm$7.2} & 39.1{$\pm$3.5} & 33.5{$\pm$1.9} & 64.2{$\pm$0.9} & 63.3{$\pm$5.2} \\
   & \citet{datafree}                     & 15.1{$\pm$6.3} & 12.9{$\pm$6.5} & 18.3{$\pm$7.2} & 20.8{$\pm$2.6} & 50.4{$\pm$3.6} & 41.6{$\pm$3.5} \\
   & \textit{SSL + Random}                & 39.5{$\pm$9.2} & 42.1{$\pm$8.6} & 42.5{$\pm$0.8} & 42.1{$\pm$2.7} & 62.0{$\pm$5.0} & 66.0{$\pm$5.8} \\
   & \textbf{SSL + Select (ours)}         & {46.0}{$\pm$6.8} & {46.2}{$\pm$6.5} & {53.0}{$\pm$2.4} & {55.2}{$\pm$3.4} & {65.1}{$\pm$2.7} & {68.5}{$\pm$5.2} \\
\midrule
\multirow{4}{*}{\textbf{GCN}} 
   & \textit{E2E} \cite{wu2021model}       & 38.2{$\pm$8.1} & 39.0{$\pm$6.1} & 28.2{$\pm$5.9} & 27.2{$\pm$3.7} & 60.1{$\pm$5.2} & 56.8{$\pm$6.1} \\
   & \citet{datafree}                     & 23.0{$\pm$4.6} & 22.5{$\pm$9.5} & 17.5{$\pm$10.5} & 16.0{$\pm$7.0} & 48.5{$\pm$1.5} & 39.0{$\pm$1.1} \\
   & \textit{SSL + Random}                & 54.7{$\pm$5.5} & 56.3{$\pm$6.6} & 51.6{$\pm$54.0} & 54.7{$\pm$3.7} & 57.6{$\pm$1.0} & 60.8{$\pm$3.8} \\
   & \textbf{SSL + Select (ours)}         & \textbf{64.3}{$\pm$3.2} & \textbf{69.6}{$\pm$2.0} & \textbf{67.3}{$\pm$2.7} & \textbf{73.2}{$\pm$3.5} & \textbf{61.9}{$\pm$2.1} & \textbf{67.1}{$\pm$3.3} \\
\bottomrule
\end{tabular}
\end{table*}
\begin{figure}[t!]%
    \centering
    \includegraphics[width=0.6\textwidth]{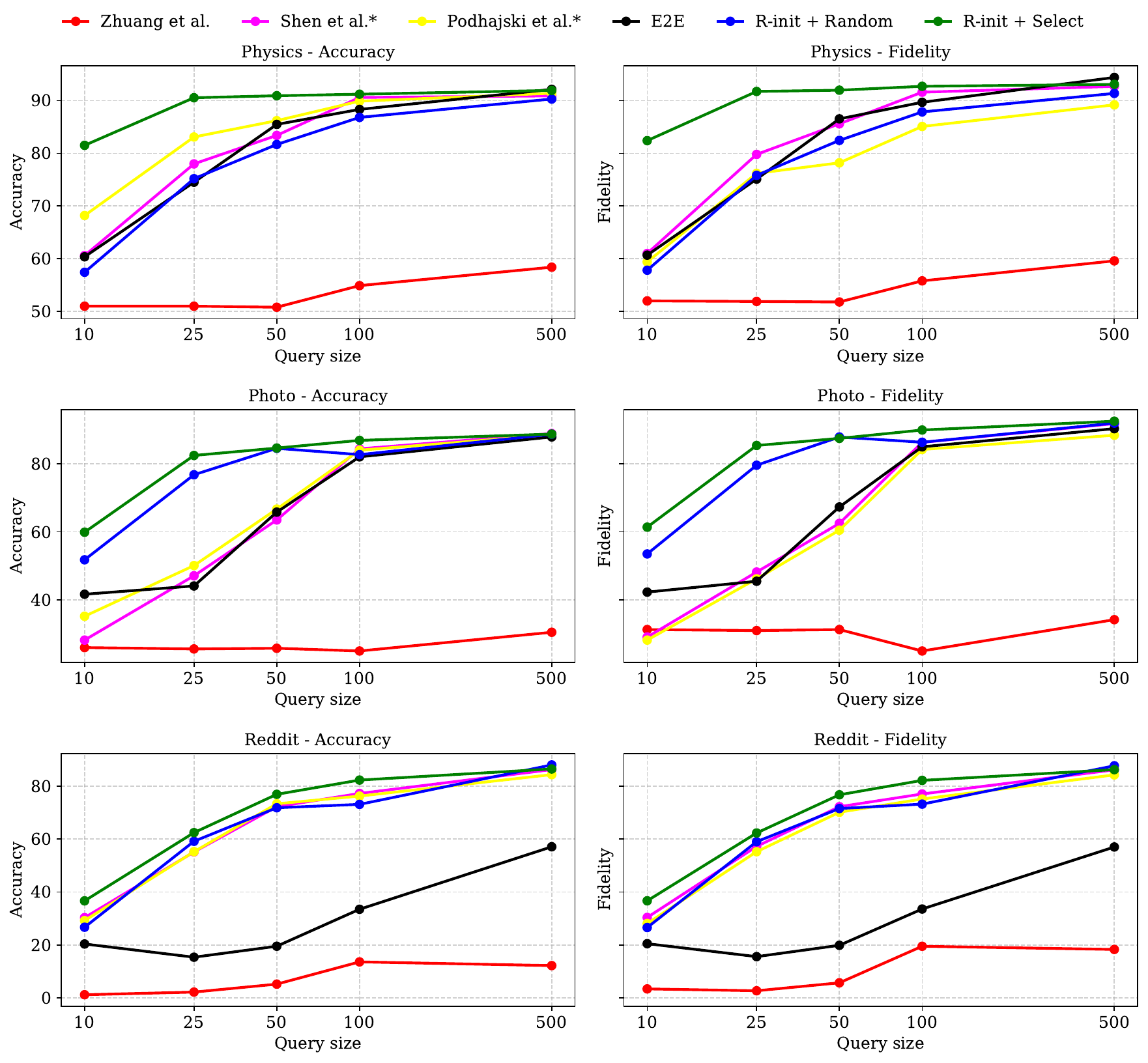}
\caption{
Accuracy and fidelity (inductive setting, target: SAGE, surrogate: GCN) for $q_n \in \{10, 25, 50, 100, 500\}$ on Physics, Photo, and Reddit. 
Methods marked with * assume access to victim embeddings (weaker threat model).
}

    \label{fig:grp1}
\end{figure}

\begin{figure*}[h!]%
    \centering
    \includegraphics[width=0.59\textwidth]{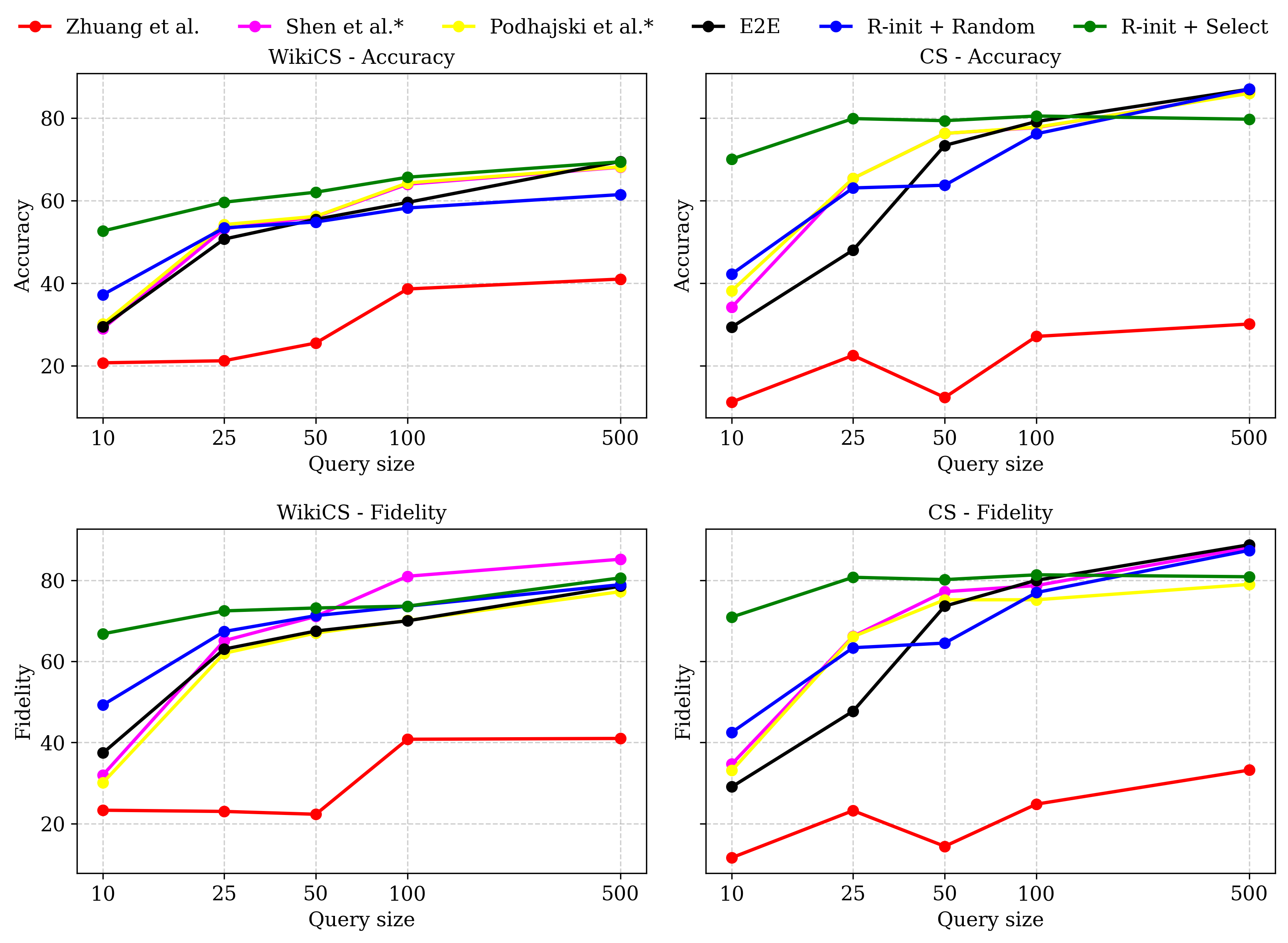}
\caption{Accuracy and fidelity (inductive setting, target: SAGE, surrogate: GCN) for $q_n \in \{10, 25, 50, 100, 500\}$ on WikiCS, and CS. 
Methods marked with * assume access to victim embeddings (weaker threat model).
}
    \label{fig:grp2}
\end{figure*}
\begin{figure*}[h!]%
    \centering
    \includegraphics[width=0.81\textwidth]{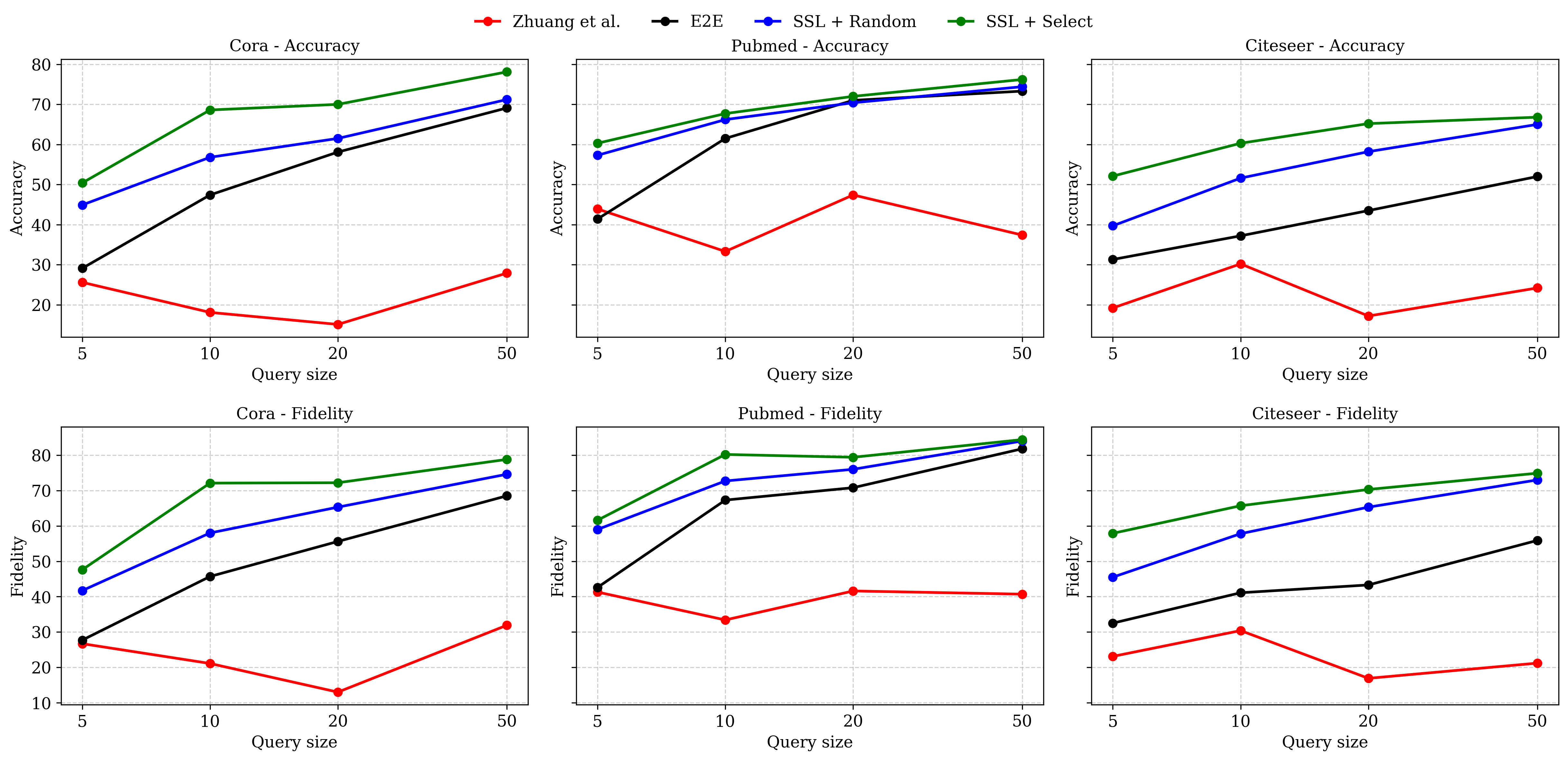}
\caption{Accuracy and fidelity (transductive setting, target: GCN, surrogate: GCN) for $q_n \in \{5,10,20,50\}$ on Cora, Pubmed, and Citeseer. 
Methods marked with * assume access to victim embeddings (weaker threat model).
}   
    \label{fig:grp3}
\end{figure*}

\begin{table}[h!]
\centering
\small
\caption{Inductive setting (target: SAGE, surrogate: GCN, $q_n = 100$, 70 test samples). P-values from McNemar’s test reported as mean ± std.\ dev.\ over 3 runs. Higher indicates greater similarity between the original and stolen models.}
\label{tab:pvalues}
\setlength{\tabcolsep}{3pt}
\begin{tabular}{l 
                c c c c c}
\toprule
\textbf{Method} & 
\textbf{Reddit} & 
\textbf{CS} & 
\textbf{Physics} & 
\textbf{Photo} & 
\textbf{WikiCS} \\
\midrule
\textit{E2E} & 
0.00{$\pm$0.00} & 0.05{$\pm$0.01} & 0.04{$\pm$0.01} & 0.02{$\pm$0.01} & 0.02{$\pm$0.01} \\

\textit{R-init + R} & 
0.08{$\pm$0.00} & 0.07{$\pm$0.01} & 0.07{$\pm$0.01} & 0.12{$\pm$0.02} & 0.05{$\pm$0.01} \\

\citet{shen2021model} & 
0.08{$\pm$0.02} & 0.01{$\pm$0.01} & 0.04{$\pm$0.01} & 0.09{$\pm$0.01} & 0.04{$\pm$0.02} \\

\citet{Podhajski_2024} & 
0.08{$\pm$0.02} & 0.01{$\pm$0.01} & 0.07{$\pm$0.01} & 0.10{$\pm$0.02} & 0.03{$\pm$0.01} \\

\citet{datafree} & 
0.00{$\pm$0.00} & 0.00{$\pm$0.00} & 0.00{$\pm$0.00} & 0.00{$\pm$0.00} & 0.00{$\pm$0.00} \\

\textbf{R-init + S (ours)} & 
\textbf{0.10}{$\pm$0.01} & \textbf{0.10}{$\pm$0.01} & \textbf{0.30}{$\pm$0.02} & \textbf{0.16}{$\pm$0.02} & \textbf{0.09}{$\pm$0.02} \\
\bottomrule
\end{tabular}
\end{table}

\begin{table}[h!]
\centering
\small
\caption{Transductive setting (target: GCN, surrogate: GCN, $q_n = 10$, 70 test samples). P-values from McNemar’s test reported as mean ± std.\ dev.\ over 3 runs. Higher indicates greater similarity between the original and stolen models.}
\label{tab:pvalues_trans}
\setlength{\tabcolsep}{3pt}
\begin{tabular}{l 
                c c c}
\toprule
\textbf{Method} &
\textbf{Cora} & 
\textbf{Citeseer} & 
\textbf{Pubmed} \\
\midrule
\textit{E2E} & 
0.00{$\pm$0.00} & 0.00{$\pm$0.00} & 0.12{$\pm$0.06} \\

\textit{SSL + Random} & 
0.02{$\pm$0.01} & 0.12{$\pm$0.06} & 0.08{$\pm$0.04} \\

\citet{datafree} & 
0.00{$\pm$0.00} & 0.00{$\pm$0.00} & 0.00{$\pm$0.00} \\

\textbf{SSL + Select (ours)} & 
\textbf{0.31}{$\pm$0.06} & \textbf{0.39}{$\pm$0.09} & \textbf{0.22}{$\pm$0.07} \\
\bottomrule
\end{tabular}

\end{table}

\begin{table}[t]
\centering
\small
\caption{
Performance and efficiency in the inductive setting ($q_n{=}100$, target: SAGE). Accuracy (Acc.) and Fidelity (Fid.) are reported as mean ± std.\ dev.\ over 3 runs. Relative time is reported with respect to our method (1.0$\times$). * indicates access to victim embeddings (weaker threat model).
}
\label{tab:gin_sage_gat}

\setlength{\tabcolsep}{3pt}
\begin{tabular}{l
                ccc
                ccc}
\toprule
\multirow{2}{*}{\textbf{Method}} &
\multicolumn{3}{c}{\textbf{Photo}} &
\multicolumn{3}{c}{\textbf{Physics}} \\
\cmidrule(lr){2-4} \cmidrule(lr){5-7}
& \textbf{Acc.} & \textbf{Fid.} & \textbf{Rel. Time} 
& \textbf{Acc.} & \textbf{Fid.} & \textbf{Rel. Time} \\
\midrule
Target accuracy & 93.0 &  &  & 96.0 &  & \\
\midrule
\textit{GIN} (\citeauthor{shen2021model})*    & 71.0*{$\pm$2.8} & 73.2{$\pm$3.2} & 5.2$\times$ & 87.4*{$\pm$0.3} & 89.1{$\pm$0.6} & 5.3$\times$ \\
\textit{GAT} (\citeauthor{shen2021model})*    & 73.4*{$\pm$1.9} & 74.9{$\pm$2.2} & 7.3$\times$ & 86.9*{$\pm$1.9} & 89.1{$\pm$1.7} & 16.2$\times$ \\
\textit{SAGE} (\citeauthor{shen2021model})*   & 68.1*{$\pm$1.1} & 70.6{$\pm$0.8} & 5.3$\times$ & 88.0*{$\pm$0.2} & 89.6{$\pm$0.2} & 7.9$\times$ \\
\textit{GIN} (\citeauthor{Podhajski_2024})*   & 72.5*{$\pm$4.4} & 72.2{$\pm$4.2} & 40.3$\times$ & 88.1*{$\pm$0.6} & 89.5{$\pm$0.5} & 14.4$\times$ \\
\textit{GAT} (\citeauthor{Podhajski_2024})*   & 73.0*{$\pm$2.8} & 74.9{$\pm$3.9} & 45.6$\times$ & 87.4*{$\pm$0.5} & 88.8{$\pm$0.5} & 15.7$\times$ \\
\textit{SAGE} (\citeauthor{Podhajski_2024})*  & 67.8*{$\pm$4.2} & 70.2{$\pm$5.3} & 44.6$\times$ & 88.2*{$\pm$0.8} & 89.9{$\pm$0.6} & 14.5$\times$ \\
\textbf{R-init + Select (ours)}              & \textbf{86.8}{$\pm$1.0} & \textbf{89.8}{$\pm$0.9} & \textbf{1$\times$22.7s{$\pm$0.8s}} & 
\textbf{91.2}{$\pm$0.4} & \textbf{92.7}{$\pm$0.5} & \textbf{1$\times$6.1s{$\pm$0.2s}} \\
\bottomrule
\end{tabular}
\end{table}

\begin{figure}[h!]
    \centering
    \begin{minipage}[t]{0.48\textwidth}
        \centering
        \includegraphics[scale=0.45]{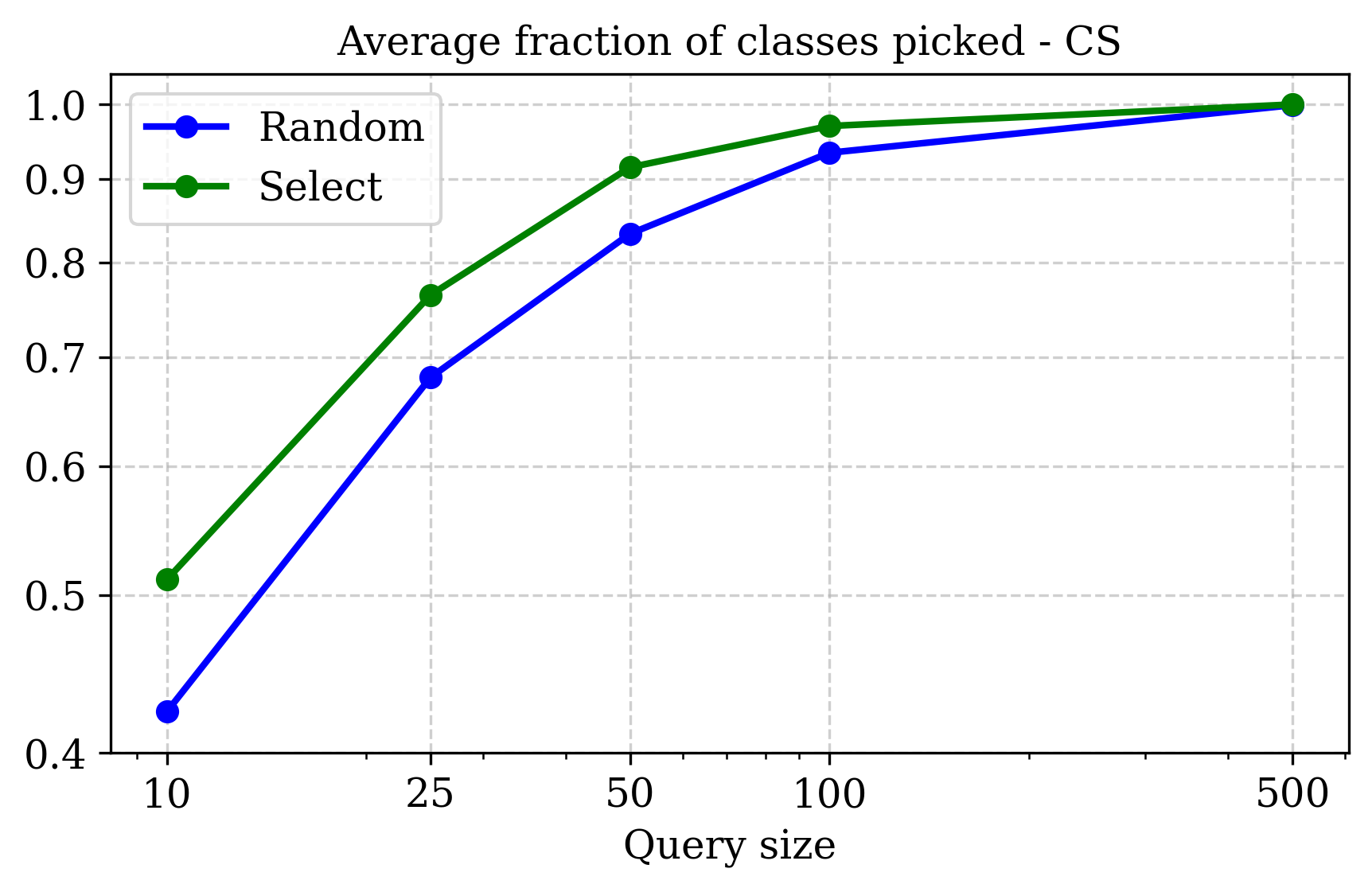}
        \caption{Average fraction of classes picked at least once using queries selected randomly vs. those selected by our method. Results are reported for the inductive setup on the CS dataset, using a SAGE target model and a GCN surrogate model. Averages are computed over 100 runs.}
        \label{fig:random_vs_select}
    \end{minipage}%
    \hfill
    \begin{minipage}[t]{0.48\textwidth}
        \centering
        \includegraphics[scale=0.23, trim=1cm 0cm 0cm 0cm, clip]{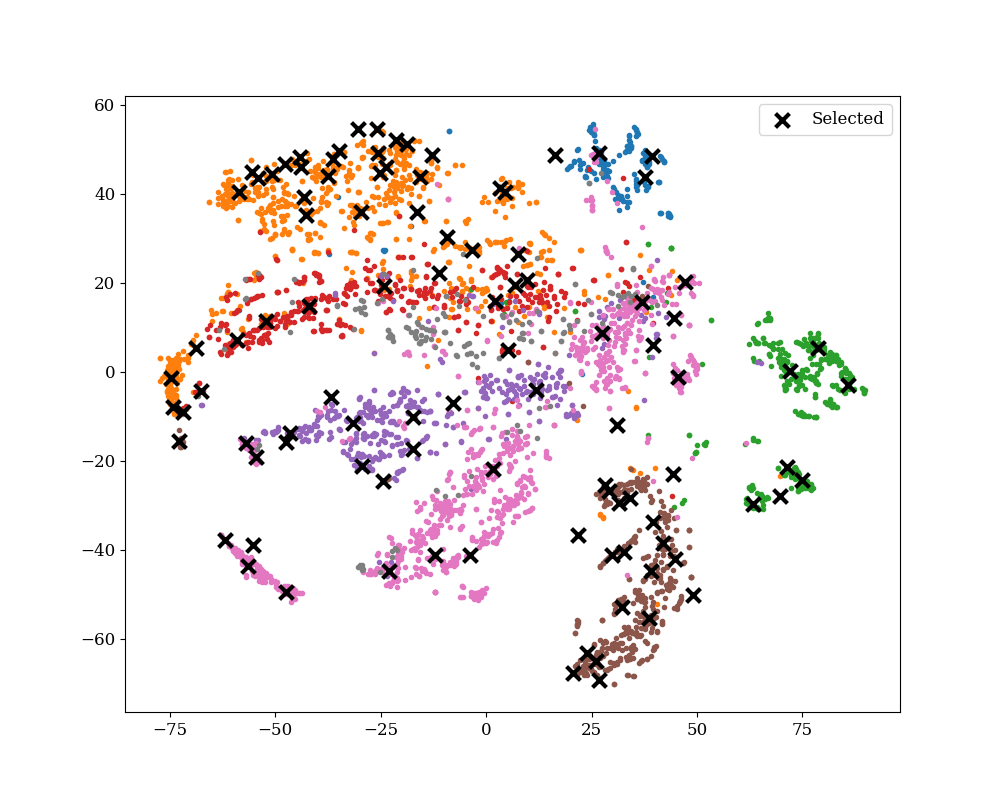}
        \caption{Visualization of selected query points on T-SNE embedding projections using the Photo dataset and $q_n=100$.}
        \label{fig:selected}
    \end{minipage}
\end{figure}

\begin{table}[h!]
\centering
\small
\caption{Comparison of selection methods for Physics, CS, and Cora datasets. }
\label{tab:active_learning_results}
\setlength{\tabcolsep}{3pt}
\begin{tabular}{l 
                cc cc cc}
\toprule
\textbf{Method} &
\multicolumn{2}{c}{\textbf{Physics}} &
\multicolumn{2}{c}{\textbf{CS}} &
\multicolumn{2}{c}{\textbf{Cora}} \\
\cmidrule(lr){2-3} \cmidrule(lr){4-5} \cmidrule(lr){6-7}
 & \textbf{Acc.} & \textbf{Fid.} & \textbf{Acc.} & \textbf{Fid.} & \textbf{Acc.} & \textbf{Fid.} \\
\midrule
Farthest-First &
77.0{$\pm$4.3} & 77.2{$\pm$4.6} &
54.5{$\pm$0.2} & 55.0{$\pm$0.2} &
60.5{$\pm$7.4} & 62.3{$\pm$7.1} \\

K-Center Greedy &
77.3{$\pm$4.4} & 77.8{$\pm$4.5} &
54.5{$\pm$0.2} & 55.0{$\pm$0.2} &
60.4{$\pm$7.4} & 62.4{$\pm$7.0} \\

Entropy Sampling &
56.4{$\pm$2.3} & 54.1{$\pm$2.0} &
57.9{$\pm$0.1} & 58.1{$\pm$0.1} &
31.8{$\pm$7.1} & 34.6{$\pm$7.9} \\

Coreset Herding &
88.4{$\pm$0.7} & 87.8{$\pm$0.5} &
76.1{$\pm$0.1} & 76.8{$\pm$0.0} &
63.4{$\pm$5.4} & 65.4{$\pm$4.1} \\

Margin Sampling &
82.7{$\pm$0.9} & 83.4{$\pm$0.8} &
75.4{$\pm$0.3} & 76.1{$\pm$0.7} &
56.5{$\pm$8.6} & 54.0{$\pm$9.7} \\

\textbf{K-Means (ours)} &
\textbf{91.2}{$\pm$0.4} & \textbf{92.7}{$\pm$0.5} &
\textbf{78.4}{$\pm$2.1} & \textbf{79.2}{$\pm$2.2} &
\textbf{69.9}{$\pm$1.2} & \textbf{72.5}{$\pm$1.3} \\
\bottomrule
\end{tabular}
\end{table}

\begin{table}[h!]
\centering
\small
\caption{Ablation study on the Physics dataset under distribution shift (node and edge perturbations). Accuracy (Acc.) and Fidelity (Fid.) are reported as mean ± std.\ dev.\ in percentage over 3 runs.}
\label{tab:physics_ablation}
\setlength{\tabcolsep}{4pt}
\begin{tabular}{lccc}
\toprule
\textbf{Change} & \textbf{Victim Acc.} & \textbf{Acc.} & \textbf{Fid.} \\
\midrule
Original & 0.96 & 90.0{$\pm$0.9} & 84.1{$\pm$0.8} \\
\cmidrule(lr){1-4}

\multicolumn{4}{l}{\textit{Node feature change}} \\
\cmidrule(lr){1-4}
5\% & 0.96 & 90.0{$\pm$0.8} & 84.1{$\pm$0.8} \\
10\% & 0.95 & 89.6{$\pm$0.9} & 83.9{$\pm$0.9} \\
30\% & 0.95 & 89.5{$\pm$0.6} & 83.8{$\pm$0.9} \\
50\% & 0.94 & 88.5{$\pm$0.7} & 83.9{$\pm$0.7} \\
70\% & 0.92 & 82.8{$\pm$2.0} & 80.8{$\pm$5.0} \\
\midrule
\multicolumn{4}{l}{\textit{Edge dropout}} \\
\cmidrule(lr){1-4}
30\% & 0.95 & 88.5{$\pm$0.5} & 89.6{$\pm$0.5} \\
50\% & 0.95 & 88.2{$\pm$1.0} & 89.1{$\pm$1.1} \\
90\% & 0.94 & 81.5{$\pm$1.3} & 81.7{$\pm$3.0} \\
100\% & 0.95 & 77.3{$\pm$2.1} & 77.3{$\pm$2.1} \\
\bottomrule
\end{tabular}
\end{table}

\clearpage

\subsection{Defense Evaluation Results}
\begin{table*}[h!]
\centering
\small
\caption{Inductive setting with label-flipping defense ($q_n = 100$). The victim (SAGE) has 10\% random label noise; the surrogate is a GCN. Accuracy (Acc.) and Fidelity (Fid.) are reported as mean ± std.\ dev.\ in percentage over 3 runs. Methods marked with * assume access to victim embeddings (weaker threat model).}
\label{tab:defense_ind}
\setlength{\tabcolsep}{1pt}
\begin{tabular}{l 
                cc cc cc cc cc}
\toprule
\textbf{Method} &
\multicolumn{2}{c}{\textbf{Reddit}} &
\multicolumn{2}{c}{\textbf{CS}} &
\multicolumn{2}{c}{\textbf{Physics}} &
\multicolumn{2}{c}{\textbf{Photo}} &
\multicolumn{2}{c}{\textbf{WikiCS}} \\
\cmidrule(lr){2-3} \cmidrule(lr){4-5} \cmidrule(lr){6-7} \cmidrule(lr){8-9} \cmidrule(lr){10-11}
 & \textbf{Acc.} & \textbf{Fid.} & \textbf{Acc.} & \textbf{Fid.} & \textbf{Acc.} & \textbf{Fid.} & \textbf{Acc.} & \textbf{Fid.} & \textbf{Acc.} & \textbf{Fid.} \\
\midrule
Target accuracy & 94.8 &  & 93.9 &  & 96.0 &  & 93.0 &  & 72.5 & \\
\cmidrule(lr){2-11}
\textit{E2E} & 
28.6{$\pm$0.1} & 26.2{$\pm$0.1} & 
74.0{$\pm$0.2} & 68.5{$\pm$0.2} & 
89.0{$\pm$0.4} & 83.1{$\pm$0.5} & 
68.1{$\pm$2.4} & 64.4{$\pm$1.8} & 
60.8{$\pm$0.1} & 62.9{$\pm$0.8} \\

\textit{R-init + Random} & 
70.1{$\pm$5.4} & 62.9{$\pm$4.6} & 
75.7{$\pm$1.2} & 69.8{$\pm$1.2} & 
83.9{$\pm$2.7} & 78.2{$\pm$2.6} & 
81.2{$\pm$2.6} & 76.9{$\pm$1.6} & 
59.0{$\pm$1.1} & 69.3{$\pm$1.1} \\
\cmidrule(lr){2-11}
\citet{shen2021model}* & 
72.0*{$\pm$1.1} & 65.9*{$\pm$1.0} & 
72.9*{$\pm$2.1} & 67.3*{$\pm$1.5} & 
84.9*{$\pm$3.4} & 79.6*{$\pm$3.2} & 
78.3*{$\pm$3.5} & 73.9*{$\pm$3.4} & 
59.4*{$\pm$1.5} & 70.1*{$\pm$2.3} \\

\citet{Podhajski_2024}* & 
72.8*{$\pm$2.8} & 65.3*{$\pm$2.6} & 
70.3*{$\pm$2.4} & 64.6*{$\pm$2.2} & 
84.6*{$\pm$2.2} & 79.0*{$\pm$2.3} & 
78.3*{$\pm$3.2} & 74.0*{$\pm$3.0} & 
57.9*{$\pm$2.6} & 68.2*{$\pm$3.8} \\

\citet{datafree} & 
12.6{$\pm$5.1} & 16.3{$\pm$5.5} & 
11.8{$\pm$4.5} & 11.8{$\pm$4.5} & 
51.8{$\pm$6.6} & 50.9{$\pm$7.2} & 
11.1{$\pm$4.4} & 11.7{$\pm$4.9} & 
32.6{$\pm$5.1} & 30.8{$\pm$7.0} \\

\textbf{R-init + Select (ours)} & 
\textbf{79.1}{$\pm$1.7} & \textbf{71.2}{$\pm$1.6} & 
\textbf{77.9}{$\pm$1.9} & \textbf{71.5}{$\pm$1.8} & 
\textbf{90.0}{$\pm$0.9} & \textbf{84.1}{$\pm$0.8} & 
\textbf{82.5}{$\pm$1.9} & \textbf{78.0}{$\pm$2.5} & 
\textbf{62.7}{$\pm$0.8} & \textbf{73.4}{$\pm$0.7} \\
\bottomrule
\end{tabular}

\end{table*}

\begin{table*}[h!]
\centering
\small
\caption{Transductive setting with label-flipping defense ($q_n = 10$). The victim and surrogate models are both GCNs. Accuracy (Acc.) and Fidelity (Fid.) are reported as mean ± std.\ dev.\ in percentage over 3 runs. Methods marked with * assume access to victim embeddings.}
\label{tab:defense_trans}
\setlength{\tabcolsep}{3pt}
\begin{tabular}{l 
                cc 
                cc 
                cc}
\toprule
\textbf{Method} & 
\multicolumn{2}{c}{\textbf{Cora}} & 
\multicolumn{2}{c}{\textbf{Citeseer}} & 
\multicolumn{2}{c}{\textbf{Pubmed}} \\
\cmidrule(lr){2-3} \cmidrule(lr){4-5} \cmidrule(lr){6-7}
 & \textbf{Acc.} & \textbf{Fid.} & \textbf{Acc.} & \textbf{Fid.} & \textbf{Acc.} & \textbf{Fid.} \\
\midrule
Target accuracy & 83.3 &  & 72.1 &  & 80.0 & \\
\cmidrule(lr){2-7}
\textit{E2E} \cite{wu2021model} & 
40.1{$\pm$8.4} & 32.1{$\pm$7.7} & 
33.9{$\pm$8.4} & 29.9{$\pm$7.6} & 
56.0{$\pm$4.9} & 53.9{$\pm$8.3} \\

\citet{datafree} & 
11.6{$\pm$1.6} & 11.0{$\pm$2.3} & 
17.2{$\pm$3.9} & 16.4{$\pm$6.6} & 
30.5{$\pm$4.0} & 31.1{$\pm$7.3} \\

\textit{SSL + Random} & 
42.2{$\pm$12.1} & 41.0{$\pm$10.5} & 
54.2{$\pm$18.1} & 52.7{$\pm$16.0} & 
55.0{$\pm$4.3} & 54.2{$\pm$3.2} \\

\textbf{SSL + Select (ours)} & 
\textbf{61.5}{$\pm$11.4} & \textbf{60.4}{$\pm$9.9} & 
\textbf{62.7}{$\pm$5.7} & \textbf{62.6}{$\pm$6.5} & 
\textbf{57.2}{$\pm$7.3} & \textbf{58.1}{$\pm$9.0} \\
\bottomrule
\end{tabular}

\end{table*}

\fi
\end{document}